\documentclass[lettersize,journal]{IEEEtran}
\usepackage{amsmath,amsfonts}
\usepackage{algorithmic}
\usepackage{algorithm}
\usepackage{array}
\usepackage[caption=false,font=normalsize,labelfont=sf,textfont=sf]{subfig}
\usepackage{textcomp}
\usepackage{stfloats}
\usepackage{url}
\usepackage{verbatim}
\usepackage{graphicx}
\usepackage{cite}

\usepackage{multirow}
\usepackage{booktabs}
\usepackage{hyperref}
\usepackage{subcaption}
\usepackage{makecell}
\usepackage{xcolor}
\definecolor{BLUE}{rgb}{0,0,1} 
\usepackage{colortbl}

\begin{document}

\title{CMTA: Leveraging Cross-Modal Temporal Artifacts for Generalizable AI-Generated Video Detection}

\author{Hang~Wang,~\IEEEmembership{Member,~IEEE,}
        Chao~Shen,~\IEEEmembership{Fellow,~IEEE,}
        Chenhao~Lin,~\IEEEmembership{Member,~IEEE,}
        Minghui~Yang,
        Lei~Zhang,~\IEEEmembership{Fellow,~IEEE,}
        and~Cong~Wang,~\IEEEmembership{Fellow,~IEEE}
\thanks{H. Wang is with Xi’an Jiaotong University, Xi'an, China, and The Hong Kong Polytechnic University, Hong Kong, China (e-mail: cshangwang@xjtu.edu.cn).}
\thanks{C. Shen and C. Lin are with Xi’an Jiaotong University, Xi'an, China (e-mail: chaoshen@mail.xjtu.edu.cn, linchenhao@xjtu.edu.cn). (Corresponding author: Chao Shen.)}
\thanks{M. Yang is with Guangdong OPPO Mobile Communications Co., Ltd.  (e-mail: yangminghui@oppo.com)}
\thanks{L. Zhang is with The Hong Kong Polytechnic University, Hong Kong, China (e-mail: cslzhang@comp.polyu.edu.hk).}
\thanks{C. Wang is with City University of Hong Kong, Hong Kong, China (e-mail: congwang@cityu.edu.hk).}
}

\markboth{Journal of \LaTeX\ Class Files,~Vol.~14, No.~8, August~2021}%
{Shell \MakeLowercase{\textit{et al.}}: A Sample Article Using IEEEtran.cls for IEEE Journals}


\maketitle

\begin{abstract}
    The proliferation of advanced AI video synthesis techniques poses an unprecedented challenge to digital video authenticity.
    Existing AI-generated video (AIGV) detection methods primarily focus on uni-modal or spatiotemporal artifacts, but they overlook the rich cues within the visual-textual cross-modal space, especially the temporal stability of semantic alignment.
    In this work, we identify a distinctive fingerprint in AIGVs, termed cross-modal temporal artifact (CMTA).
    Unlike real videos that exhibit natural temporal fluctuations in cross-modal alignment due to semantic variations, AIGVs display unnaturally stable semantic trajectories governed by given input prompts.
    To bridge this gap, we propose the CMTA framework, a cross-modal detection approach that captures these unique temporal artifacts through joint cross-modal embedding and multi-grained temporal modeling. 
    Specifically, CMTA leverages BLIP to generate frame-level image captions and utilizes CLIP to extract corresponding visual-textual representations.
    A coarse-grained temporal modeling branch is then designed to characterize temporal fluctuations in cross-modal alignment with a GRU. 
    In parallel, a fine-grained branch is constructed to capture intricate inter-frame variations from integrated visual-textual features with a Transformer encoder.
    Extensive experiments on 40 subsets across four large-scale datasets, including GenVideo, EvalCrafter, VideoPhy, and VidProM, validate that our approach sets a new state-of-the-art while exhibiting superior cross-generator generalization.  Code and models of CMTA will be released. \footnote{~\url{https://github.com/hwang-cs-ime/CMTA}}
\end{abstract}


\section{Introduction}
\IEEEPARstart{C}{ontemporary} AI video generation models such as Veo~\cite{Veo}, Sora~\cite{brooks2024video}, and Gen-2~\cite{germanidis2023gen} typically produce cinematic-quality videos with unprecedented realism, spanning diverse scenarios from daily life scenes to specialized fields, which are indistinguishable from real videos to the human eye. 
However, this remarkable technological advancement has also introduced significant security risks that permeate societal, national, and individual levels. 
Malicious actors can utilize these AI-generated videos to disseminate disinformation, manipulate public opinion, fabricate celebrity scandals, and even create fake military and political footage, which poses severe threats to social trust, public order, and national security. 
As the quality of AI-generated videos continues to improve and their creation tools become increasingly accessible, developing highly generalizable detection methods has become crucial to safeguarding information authenticity and mitigating potential harms.

\begin{figure}[t]
\centering
\includegraphics[width=\columnwidth]{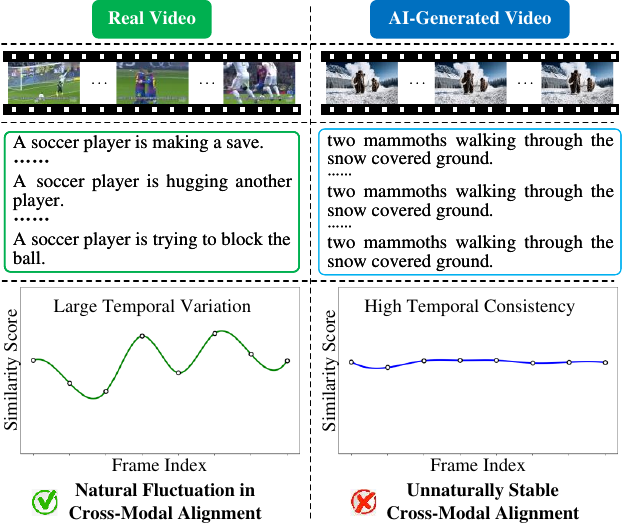} 
    \caption{Motivation of the proposed CMTA framework.
    (Left) Real videos in natural scenes exhibit large variations in high-level visual semantics and generated textual descriptions, leading to stochastic temporal fluctuations in cross-modal similarity. 
    (Right) AI-generated videos, despite appearing visually dynamic to human eyes, are synthesized under fixed semantic prompts. Their high-level visual semantics and corresponding captions remain highly consistent across frames, yielding unnaturally stable cross-modal alignment as a distinctive temporal artifact of AI-generated videos.}
\label{fig1_CMTA}
\end{figure}

Existing AI-generated video detection methods~\cite{bai2024ai, chen2024DeMamba, Ma2024DetectingAV, Zheng_2025_ICCV, zhang2025NSGVD, interno2025aigenerated} mainly rely on either modeling temporal and spatial dynamics or identifying violations of physical laws to distinguish real videos from synthetic ones.
Spatial and temporal modeling approaches~\cite{bai2024ai, chen2024DeMamba, Ma2024DetectingAV} aim to extract discriminative forensic artifacts, evolving from frame-wise anomaly integration to long-range sequential modeling and inter-frame consistency patterns.
Alternatively, theory-driven detection paradigms~\cite{Zheng_2025_ICCV, zhang2025NSGVD, interno2025aigenerated} leverage intrinsic physical or geometric priors, such as second-order dynamical discrepancies, probability flow conservation, and temporal representation curvature, to expose synthetic artifacts with minimal supervision.
Despite such advancements, existing methods primarily focus on visual-domain artifacts or low-level spatiotemporal cues, overlooking the high-level semantic constraints inherent in AI-generated videos. 
Consequently, they fail to capture the intrinsic pattern of unnaturally stable global semantic alignment that uniquely characterizes AI-generated sequences.

To bridge this gap, we investigate the inherent differences in the temporal evolution of cross-modal semantic similarity between real and AI-generated videos, as illustrated in Fig.~\ref{fig1_CMTA}.
Notably, real-world videos demonstrate natural temporal fluctuations in visual-textual semantic similarity due to dynamic scene variations, whereas AI-generated videos synthesized under fixed semantic constraints maintain abnormally stable cross-modal alignment across frames.
This intrinsic pattern constitutes a discriminative cross-modal temporal artifact that distinguishes synthetic sequences from real ones.
Motivated by this observation, we propose the CMTA framework, which explicitly models the temporal dynamics of cross-modal semantic alignment to capture these unique artifacts for accurate and generalizable video forensics.

Specifically, CMTA first leverages an image caption model to generate frame-level textual descriptions, and then adopts the CLIP visual and textual encoders to extract the corresponding visual and textual representations.
To characterize the coarse-grained temporal fluctuations of cross-modal semantic similarity, a GRU is utilized to model the temporal evolution of inter-frame alignment variations throughout the entire video sequence.
In parallel, the visual and textual features of each frame are concatenated along the channel dimension, and a Transformer encoder is employed to capture the fine-grained temporal variations of inter-frame cross-modal semantics.
By fusing the coarse-grained and fine-grained cross-modal semantic temporal representations, CMTA effectively integrates multi-granularity cross-modal temporal clues to accurately identify the unique artifacts inherent in AI-generated videos, thus enabling generalizable video forensic detection.
Ultimately, the fused joint representations are fed into a MLP classifier to predict the authenticity of the input videos.

To conclude, the main contributions are summarized as follows:
\begin{itemize}
    \item We investigate the unique cross-modal temporal artifact inherent in AI-generated videos, which has been largely overlooked by existing detection methods. Specifically, AI-generated videos exhibit abnormally stable visual-textual semantic alignment across frames, whereas real videos present natural temporal fluctuations in cross-modal semantic similarity.

    \item We propose a novel framework termed CMTA, which leverages cross-modal temporal artifacts to detect AI-generated videos. Specifically, we utilize an image caption model and CLIP encoders to obtain frame-level visual-textual representations, and then employ a GRU and Transformer encoder to model coarse- and fine-grained cross-modal temporal dynamics for capturing the underlying artifacts.

    \item Extensive evaluations on four large-scale benchmarks demonstrate that CMTA achieves state-of-the-art performance and significantly outperforms existing baselines by a notable margin. 
    Specifically, our approach attains mean AP and AUC of 98.74\% and 99.10\% on GenVideo, 99.73\% and 99.73\% on EvalCrafter, 95.54\% and 97.34\% on VideoPhy, and 94.35\% and 95.86\% on VidProM.
\end{itemize}

\begin{figure*}[t]
\centering
\includegraphics[width=0.98\textwidth]{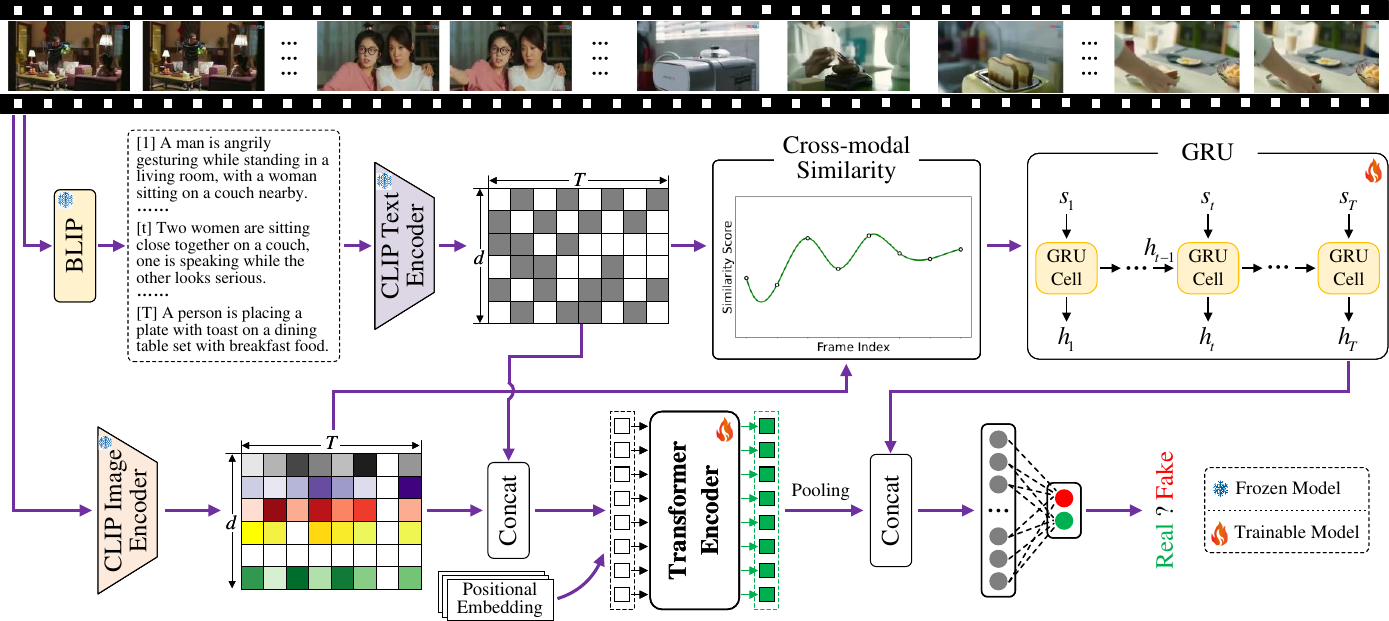} 
\caption{The pipeline of the proposed CMTA framework. Given an input video, CMTA first leverages BLIP to generate frame-wise captions, and then extract visual and textual representations via CLIP encoders. A GRU is subsequently utilized to model coarse-grained temporal dynamics, capturing the evolution of cross-modal alignment. Simultaneously, a Transformer encoder encapsulates fine-grained temporal variations based on fused cross-modal features. Finally, these multi-grained representations are aggregated and fed into a classification head for AI-generated video detection.}
\label{fig2_CMTA}
\end{figure*}

\section{Related Work}
\label{sec:relwork}

\subsection{Deepfake Video Detection}
Existing video-based deepfake detection methods can be broadly categorized into two main directions: explicit spatial-temporal artifact modeling and generalization-driven learning paradigms.

The former category centers on explicit spatial-temporal artifact capture and architectural optimization, addressing ``what to detect'' by identifying unnatural inter-frame dynamics, pixel flickering, and structural distortions through refined network architectures or novel low-level physical forensic features.
To capture diverse spatial-temporal inconsistencies, FTCN~\cite{zheng2021exploring} constrains 3D convolutions to prioritize modeling temporal inconsistencies, whereas STIL~\cite{10.1145/3474085.3475508} and DIP~\cite{nie2024dip} leverage difference modules and directional inconsistency patterns to better characterize fine-grained forgery artifacts.
MRE-Net~\cite{pang2023mre} introduces multi-rate excitation branches to capture dynamic spatial-temporal inconsistencies across multiple temporal scales, and Chen et al.~\cite{chen2022deepfake} combine spatial attention with texture enhancement to facilitate fine-grained forensic analysis.
Going beyond spatial or temporal domains, BSF~\cite{Kim_2025_ICCV} employs pixel-wise temporal frequency via 1-D Fourier Transform to capture subtle inconsistencies, while SLF~\cite{choi2024exploiting} detects anomalies in the evolution of style latent vectors.
To enhance robustness against real-world video compression, Chen et al.~\cite{chen2024compressed} leverage 3D spatiotemporal trajectories to capture consistent motion patterns and achieve reliable deepfake detection.
Regarding efficiency and interpretability, ISTVT~\cite{zhao2023istvt} devises an interpretable spatial-temporal Transformer with explicit artifact visualization, while TALL~\cite{xu2023tall} recasts video detection as an efficient thumbnail layout analysis task. 
Meanwhile, MINTIME~\cite{10547206} leverages dedicated attention mechanisms and tailored embeddings to handle multi-identity and scale-invariance forgery challenges.

The latter category pursues generalization-driven learning paradigms and multi-dimensional perception, addressing ``how to learn'' and mitigating overfitting to specific forgery algorithms via self-supervision, data augmentation, multi-modal fusion, and universal representation learning.
Regarding data and representation enhancement, FakeSTormer~\cite{nguyen2025vulnerability} uses self-blended videos to simulate subtle artifacts, Yan et al.~\cite{yan2025generalizing} introduce video-level blending to capture facial feature drift, and AltFreezing~\cite{wang2023altfreezing} employs an alternating freezing strategy to decouple spatial-temporal training.
NACO~\cite{zhang2024learning} focuses on learning natural consistency from only real videos in a self-supervised manner to handle unseen forgery attacks.
To further enhance robustness against unseen attacks, ID-Reveal~\cite{cozzolino2021id} exploits identity-aware temporal features to model person-specific facial motion patterns, while SFake~\cite{xie2024shaking} introduces active physical probing via smartphone vibrations to capture unique hardware-linked motion artifacts.
Furthermore, multi-modal detection methods such as AVFF~\cite{oorloff2024avff} and Feng et al.~\cite{feng2023self} identify fake videos by detecting audio-visual misalignment.
Toward universal forensics, UNITE~\cite{kundu2025towards} unifies face manipulation, background editing, and fully AI-generated content detection by leveraging generic foundation model features.

Despite these advancements, most existing methods are confined to the visual domain, overlooking high-level semantic alignment and temporal anomalies within the joint vision-language space. 
To bridge this gap, our CMTA framework exploits cross-modal temporal artifacts between visual semantics and textual descriptions, establishing a novel and generalizable forensic paradigm beyond conventional visual cues.

\subsection{AI-Generated Video Detection}
Existing AI-generated video detection methods~\cite{bai2024ai, chen2024DeMamba, Ma2024DetectingAV, Zheng_2025_ICCV, zhang2025NSGVD, interno2025aigenerated, vahdati2024beyond, chang2024matters, ji2024distinguish, liu2024turns} primarily rely on modeling temporal and spatial dynamics, integrating multi-modal expert knowledge, or identifying violations of physical laws to distinguish real videos from synthetic ones.

AIGVDet~\cite{bai2024ai} utilized a two-branch CNN to perform frame-by-frame prediction by capturing forensic anomalies in both spatial and optical-flow domains.
To better capture long-range temporal dependencies, DeMamba~\cite{chen2024DeMamba} introduced state space models (Mamba) to achieve more efficient sequential modeling.
Meanwhile, DeCoF~\cite{Ma2024DetectingAV} prioritized temporal modeling of frame consistency to mitigate the generator-specific characteristics of spatial artifacts and enhance cross-generator generalization.
Recognizing the importance of motion cues for generalized detection, DuB3D~\cite{ji2024distinguish} adopted a dual-branch 3D Swin Transformer to jointly model spatial-temporal and motion features, achieving strong performance on the large-scale GenVidDet dataset.
Beyond spatiotemporal modeling, Vahdati \textit{et al.}~\cite{vahdati2024beyond} revealed that synthetic image detectors often fail on AI-generated videos due to their distinct temporal artifacts, and further demonstrated that such video-specific artifacts remain robust against H.264 compression and can be adapted to unseen generators via few-shot tuning.
Similarly, Liu et al.~\cite{liu2024turns} identified severe domain gaps in methods for detecting diffusion-generated videos, and proposed the DIVID framework using diffusion reconstruction errors (DIRE) with a CNN+LSTM architecture to capture temporal dynamics for robust cross-generator detection.
Furthermore, Chang \textit{et al.}~\cite{chang2024matters} proposed an Ensemble-of-Experts model that fuses appearance cues from visual foundation models, motion features from optical flow, and geometric clues from monocular depth, thereby effectively identifying systematic artifacts in advanced AI-generated videos such as Sora and significantly improving cross-generator generalization.

Recent detection paradigms have shifted towards leveraging intrinsic physical or geometric priors to improve generalization with minimal supervision.
For instance, D3~\cite{Zheng_2025_ICCV} leverages second-order temporal difference features to distinguish real and synthetic videos in a training-free manner.
Similarly, NSG-VD~\cite{zhang2025NSGVD} employs the Normalized Spatiotemporal Gradient to capture subtle deviations from natural video dynamics.
Based on the `perceptual straightening' hypothesis, ReStraV~\cite{interno2025aigenerated} computes the curvature of video representations over time to expose geometric irregularities inherent in AI-generated sequences.

Collectively, despite these advances, existing methods still overlook the intrinsic semantic-level temporal evolution within the visual-textual cross-modal space.
To bridge this gap, our CMTA framework pioneers the exploitation of cross-modal temporal artifacts to achieve generalizable AI-generated video detection.

\section{Methodology}
\label{sec:method}
The objective of AI-generated video detection is to determine whether a given input video is authentic or AI-generated.
As shown in Fig.~\ref{fig2_CMTA}, the proposed CMTA framework mainly consists of four modules:
(1) \textit{Caption Generation and Visual-Textual Representation}, which employs the BLIP~\cite{li2022blip} model to produce frame-wise image captions and the CLIP~\cite{radford2021learning} model to extract the corresponding visual and textual representations for each frame;
(2) \textit{Coarse-grained Temporal Modeling}, where the inter-frame visual-textual similarity is processed by a GRU~\cite{cho-etal-2014-learning} to capture coarse-grained temporal dynamics;
(3) \textit{Fine-grained Temporal Modeling}, which utilizes a Transformer encoder~\cite{dosovitskiy2020vit} to build fine-grained cross-modal temporal interactions across frames based on concatenated visual-textual features;
and (4) \textit{Prediction Head}, where these multi-grained features are concatenated and passed through a fully-connected layer to predict the final results.
The details of each component are elaborated in the following sections.

\subsection{Caption Generation and Visual-Textual Representation}
Given a video, we first randomly sample \( T \) consecutive frames as input.
Then, we employ the BLIP model~\cite{li2022blip} to generate captions for each frame, aiming to explicitly capture its high-level semantics in natural language.
Subsequently, we adopt the CLIP~\cite{radford2021learning} visual and textual encoders to extract corresponding representations from each frame and its generated caption, respectively.
Together, these operations construct well-aligned cross-modal representations at frame level to facilitate subsequent multi-grained temporal modeling.

\subsubsection{Random Frame Sampling}
Given a video $V$, we randomly select a sequence of \( T \) consecutive frames, denoted as \( f_1, f_2, \dots, f_T \).
This sampling strategy allows us to capture the inherent temporal dynamics of videos, while also introducing adequate diversity for model training.

\subsubsection{Frame-Level Caption Generation}
For each frame \( f_t \) , we feed it into a Transformer-based image captioning model, i.e., BLIP~\cite{li2022blip}, to generate a textual caption  \( c_t \).
This operation distills the visual semantics of each frame into linguistic descriptions. 
Formally, this process can be written as:
\begin{equation}
    c_t = \text{BLIP}(f_t), \quad t \in \{1, 2, \dots, T\}
\end{equation}

\subsubsection{Visual and Textual Feature Extraction}
After obtaining the caption for each frame, we leverage the CLIP model~\cite{radford2021learning} to extract visual and textual features.
Specifically, CLIP projects images and texts into a shared semantic space, facilitating the modeling of intrinsic correlations between visual content and associated textual descriptions.
Concretely, the visual encoder of CLIP is utilized to extract the visual feature $\mathbf{v}_t$ from frame \( f_t \), while the textual encoder is employed to derive the textual feature $\mathbf{e}_t$ from the associated caption \( c_t \):
\begin{equation}
    \mathbf{v}_t = \text{CLIP}_{\text{visual}}(f_t),
\end{equation}
\begin{equation}
    \mathbf{e}_t = \text{CLIP}_{\text{textual}}(c_t),
\end{equation}
Such paired frame-wise visual-textual features encode compact, aligned semantics for every frame, which serve as the basis for the following coarse-grained and fine-grained temporal modeling.

\begin{figure}[t]
\centering
\includegraphics[width=\columnwidth]{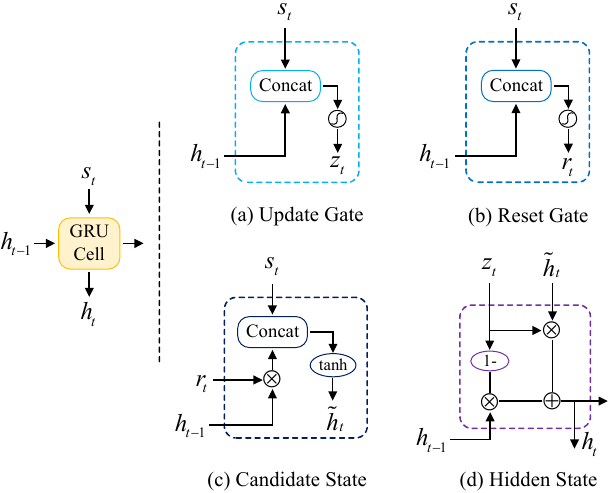} 
    \caption{Illustration of the GRU cell for coarse-grained temporal modeling. It consists of an update gate, a reset gate, a candidate hidden state, and an output hidden state. Given the cross-modal similarity sequence as input, the GRU adaptively aggregates temporal dynamics and outputs the final hidden state as the coarse-grained temporal representation.}
\label{fig3_GRU}
\end{figure}

\subsection{Coarse-Grained Temporal Modeling}
The goal of coarse-grained temporal modeling is to capture the global temporal dynamics and overall semantic trend of the video, providing a global representation for final prediction.
Based on the paired visual and textual features extracted above, we first quantify the frame-wise cross-modal semantic correlation by computing the visual-textual similarity score $s_t$.
\begin{equation}
    s_t = \frac{\mathbf{v}_t \cdot \mathbf{e}_t}{\| \mathbf{v}_t \| \|\mathbf{e}_t\|},
\end{equation}
where a larger \( s_t \) indicates a higher degree of semantic alignment between the visual frame and its caption.
Accordingly, we define the cross-modal similarity sequence across all frames as $\mathbf{S} = \{s_t\}_{t=1}^{T}$.
Notably, the temporal evolution of cross-modal alignment differs significantly between AI-generated and real videos.
Real-world videos contain natural and diverse scenes with evolving visual semantics, giving rise to intuitive and irregular fluctuations in cross-modal similarity.
Conversely, the synthesis of AI-generated videos is governed by fixed semantic constraints, leading to an unnaturally stable cross-modal alignment. 
This phenomenon manifests as a discriminative temporal artifact that effectively characterizes synthetic videos.

To capture the temporal evolution of cross-modal alignment in video $V$, we employ a Gated Recurrent Unit (GRU)~\cite{cho-etal-2014-learning} to process the similarity sequence $\mathbf{S}$. 
The detailed architecture of the GRU cell is illustrated in Fig.~\ref{fig3_GRU}.
Benefiting from its gated update mechanism, the GRU enables efficient modeling of temporal dynamics in cross-modal alignment with fewer parameters while alleviating gradient vanishing during training. 
Specifically, the operation at each time step $t$ is formulated as follows:
\begin{equation}
\begin{aligned}
\mathbf{z}_t &= \sigma\left(\mathbf{W}_z \cdot [\mathbf{h}_{t-1}, s_t]\right), \\
\mathbf{r}_t &= \sigma\left(\mathbf{W}_r \cdot [\mathbf{h}_{t-1}, s_t]\right), \\
\tilde{\mathbf{h}}_t &= \tanh\left(\mathbf{W}_h \cdot [\mathbf{r}_t \odot \mathbf{h}_{t-1}, s_t]\right), \\
\mathbf{h}_t &= (1 - \mathbf{z}_t) \odot \mathbf{h}_{t-1} + \mathbf{z}_t \odot \tilde{\mathbf{h}}_t,
\end{aligned}
\label{eq:gru_t}
\end{equation}
where $\mathbf{z}_t$ and $\mathbf{r}_t$ denote the update gate and reset gate, respectively, $\tilde{\mathbf{h}}_t$ is the candidate hidden state, $\sigma$ is the sigmoid function, and $\odot$ represents the element-wise product.
By processing the similarity sequence $\mathbf{S}$ sequentially, the GRU adaptively aggregates the temporal fluctuations inherent in frame-wise cross-modal alignment.
The final hidden state $\textbf{h}_T$ at the last time step $T$ is adopted as the coarse-grained temporal representation of the video.
This representation explicitly encodes the discriminative temporal patterns of cross-modal alignment that distinguish real and AI-generated videos.
By aggregating global temporal correlations across all frames, $\mathbf{h}_T$ captures the overall semantic evolution of the video, providing a high-level discriminative basis for final prediction.

\subsection{Fine-Grained Temporal Modeling}
Complementing the coarse-grained branch that models global trends of cross-modal similarity, the fine-grained temporal modeling branch focuses on capturing detailed inter-frame feature interactions and subtle temporal dependencies from the paired visual-textual features.
Notably, the two branches form a pure parallel and complementary structure, jointly capturing multi-perspective temporal characteristics to facilitate accurate discrimination between real and AI-generated videos.

For each frame $t$, we first construct the unified cross-modal feature by concatenating the visual feature $\mathbf{v}_t$ and textual feature $\mathbf{e}_t$, which is formulated as:
\begin{equation}
    \mathbf{x}_t = [\mathbf{v}_t; \mathbf{e}_t]
\end{equation}
where $[\cdot; \cdot]$ denotes the channel-wise concatenation operation. 
To unify the feature dimension for subsequent temporal modeling, we project $\mathbf{x}_t$ into a dense embedding space via a linear projection layer:
\begin{equation}
    \mathbf{u}_t = \mathbf{W}_p \mathbf{x}_t + \mathbf{b}_p
\end{equation}
where $\mathbf{W}_p$ and $\mathbf{b}_p$ represent the learnable weight and bias of the projection layer, respectively.

To preserve the temporal order of video frames, we introduce learnable positional embeddings tailored for the fixed frame length of our input. 
We add the positional embedding $\mathbf{P}$ to the frame feature sequence to aggregate temporal and semantic information:
\begin{equation}
    \mathbf{U}_0 = [\mathbf{u}_1; \mathbf{u}_2; \dots; \mathbf{u}_T] + \mathbf{P}
\end{equation}
where $\mathbf{P}$ is the learnable positional embedding parameter optimized during training, and $\mathbf{U}_0$ serves as the input of the Transformer encoder.

We utilize a Transformer encoder to model fine-grained temporal dependencies.
Equipped with multi-head self-attention, the encoder explicitly computes the temporal correlation between every pair of cross-modal frame features, capturing both short-range local fluctuations and long-range contextual interactions.
Based on the integrated frame feature sequence, the self-attention mechanism computes the adaptive inter-frame affinity to mine subtle temporal patterns critical for video forensics:
\begin{equation}
    \text{Attn}(\mathbf{U}_0) = \text{Softmax}\left( \frac{\mathbf{U}_0\mathbf{W}_q (\mathbf{U}_0\mathbf{W}_k)^\top}{\sqrt{d_k}} \right) (\mathbf{U}_0\mathbf{W}_v)
\end{equation}
where $d_k$ denotes the dimension of the query and key projection, and $\mathbf{W}_q$, $\mathbf{W}_k$, $\mathbf{W}_v$ are projection matrices dedicated to cross-modal frame feature interaction.

After Transformer encoding, we perform average pooling along the temporal dimension to aggregate frame-level embeddings into a compact representation:
\begin{equation}
    \mathbf{h}_{\text{fine}} = \frac{1}{T} \sum_{t=1}^{T} \mathbf{U}^L_t,
\end{equation}
where $\mathbf{U}^L_t$ denotes the encoded feature of the $t$-th frame output by the final $L$-th layer of the Transformer encoder, and $\mathbf{h}_{\text{fine}}$ represents the final fine-grained temporal representation.

This branch captures delicate inter-frame variations that provide critical clues for AI-generated video detection.
Real videos exhibit significant visual-textual variations and natural temporal fluctuations, whereas AI-generated videos produced via anchor-based generation processes present unnaturally stable cross-modal alignment as a typical temporal artifact.
By explicitly encoding such temporal artifacts, the fine-grained branch complements the global temporal patterns characterized by the coarse-grained branch.
Benefiting from the parallel dual-branch architecture, the proposed CMTA framework comprehensively captures multi-grained temporal dynamics from both global trends and inter-frame variations, enabling reliable and accurate detection of AI-generated videos.

\subsection{Prediction Head and Loss}
After obtaining the coarse-grained temporal representation $\mathbf{h}_T$ from the GRU branch and the fine-grained temporal representation $\mathbf{h}_{fine}$ from the Transformer branch, we fuse these two complementary features for the final classification. 
Specifically, the prediction head integrates the global temporal dynamics of cross-modal alignment and the subtle inter-frame temporal correlation to identify real and AI-generated videos.
We concatenate these two discriminative temporal representations to form a fused representation:
\[
\mathbf{h}_{\text{fusion}} = [\mathbf{h}_T; \mathbf{h}_{fine}].
\]
where $[\cdot; \cdot]$ represents the concatenation operation along the channel dimension.

Subsequently, the fused embedding \( \mathbf{h}_{\text{fusion}} \) is fed into a fully connected layer followed by a Softmax function to produce the final prediction:
\[
\hat{y} = \text{Softmax}(\mathbf{W} \mathbf{h}_{\text{fusion}} + \mathbf{b}),
\]
where \( \mathbf{W} \) and \( \mathbf{b} \) are the learnable parameters of the FC layer, and $\hat{y} \in [0, 1]^2$ denotes the probability distribution over the two classes.

The model is optimized using the binary cross-entropy loss:
\[
\mathcal{L}_{\text{CE}} = -[y \log(\hat{y}) + (1 - y)\log(1 - \hat{y})],
\]
where $y \in \{0, 1\}$ is the ground-truth label, with $y=1$ for AI-generated videos and $y=0$ for real videos. 
$\hat{y}_0$ and $\hat{y}_1$ denote the predicted probabilities of real and AI-generated videos, respectively.
By minimizing the loss function, the model is optimized to learn discriminative multi-grained cross-modal temporal cues, ensuring accurate detection of AI-generated videos.

\begin{table*}[th]
    \centering
    \caption{Comparison of AP (\%) between CMTA and baselines on GenVideo. \textsuperscript{\textdagger} Reproduced using the official code. \textsuperscript{\textdagger\textdagger} Reproduced from our implementation due to the unavailability of official code.}
    \resizebox{\textwidth}{!}{
    \begin{tabular}{c | c c c c c c c c c c | c}
    \toprule
    \rowcolor{gray!25}
    Method                                                         & MS        & MPS        & MV         & HotShot            & Show-1            & Gen2               & Crafter               & LaVie               & Sora                 & WS             & mean                \\
    \midrule
    \hline
    \rowcolor{gray!5}
    STIL\textsuperscript{\textdagger} \cite{10.1145/3474085.3475508}, ACM MM 2021               & 88.21             & 88.68              & 73.07              & 56.91              & 62.07             & 83.96              & 64.87                 & 65.54               & 49.86                & 63.60                  & 69.68               \\ 
    FTCN\textsuperscript{\textdagger} \cite{zheng2021exploring}, ICCV 2021            & 70.01             & 83.59              & 97.07              & 87.42              & 93.30             & 91.86              & 91.72                 & 84.16               & 44.48                & 84.46                  & 82.81               \\
    \rowcolor{gray!5}
    X-CLIP\textsuperscript{\textdagger} \cite{ni2022expanding}, ECCV 2022             & 79.84             & 87.54              & 95.53              & 90.71              & 94.54             & 88.69              & 93.50                 & 86.28               & 64.23                & \underline{88.54}                  & 89.62               \\ 
    TALL\textsuperscript{\textdagger} \cite{xu2023tall}, ICCV 2023               & 51.11             & 63.63              & 92.09              & 44.00              & 51.06             & 93.47              & 87.85                 & 59.07               & 15.82                & 64.43                  & 62.25               \\ 
    \rowcolor{gray!5} 
    FID\textsuperscript{\textdagger} \cite{NEURIPS2024_6dddcff5}, NeurIPS 2024   & 91.50 & 92.24              & 93.67              & 86.10  & 90.61 & 93.27              & 92.41                 & 83.68               & 74.95                & 82.24      & 88.07   \\ 
    NPR\textsuperscript{\textdagger} \cite{Tan_2024_CVPR}, CVPR 2024             & 84.67             & 96.53              & 96.79              & 40.17              & 21.61             & 96.35              & 97.02                 & 22.37               & 90.55    & 66.51                  & 71.26               \\
    \rowcolor{gray!5} 
    MINTIME\textsuperscript{\textdagger} \cite{10547206}, TIFS 2024                  & 79.27             & 82.03              & 89.80              & 87.68              & 89.23             & 88.26              & 87.34                 & 82.48               & 80.75                & 85.10                  & 85.19               \\     
    AIGVDet\textsuperscript{\textdagger} \cite{bai2024ai}, PRCV 2024             & 70.91             & 67.93              & 56.22              & 51.81              & 72.59             & 89.98              & 75.87                 & 88.62   & 65.70                & 64.96                  & 70.46               \\  
    \rowcolor{gray!5}
    DeMamba\textsuperscript{\textdagger} \cite{chen2024DeMamba}, arXiv 2024           & 41.96             & 97.07              & 84.64              & 67.63              & 45.07             & 96.11              & \underline{98.26}                 & 81.49               & 28.79                & 78.00                  & 71.90               \\
    DeCoF\textsuperscript{\textdagger\textdagger} \cite{Ma2024DetectingAV}, ICME 2025   & 91.18             & 91.69              & \textbf{98.68}              & 76.02              & 48.99             & 98.28              & 94.67                 & 77.79               & 55.76                & 73.68                  & 81.67               \\
    \rowcolor{gray!5}
    NSG-VD\textsuperscript{\textdagger} \cite{zhang2025NSGVD}, NeurIPS 2025        & 70.01 & 83.59 & 97.07 & 87.42 & 93.30 & 91.86 & 91.72 & 84.16 & 44.48 & 84.46 & 82.81  \\
    ReStraV\textsuperscript{\textdagger} \cite{interno2025aigenerated}, NeurIPS 2025            & \underline{95.34}             & \underline{97.22}              & \underline{97.43}              & \textbf{99.19}              & 74.52             & \underline{99.92}             & 71.28             & \underline{96.80}             & \underline{97.99}             & 73.75             & 90.34              \\
    \rowcolor{gray!5} 
    D3\textsuperscript{\textdagger} \cite{Zheng_2025_ICCV}, ICCV 2025        & 85.59             & 94.07             & 96.22             & \underline{97.09}             & \underline{95.29}             & 94.65             & 96.46             & 88.22             & 87.71             & 86.30             & \underline{92.16}         \\  
    \hline \hline
    \rowcolor{gray!25}
    \textbf{CMTA}                                                           & \textbf{99.22}     & \textbf{99.59}    & 96.09              & 96.99     & \textbf{97.65}    & \textbf{99.98}     & \textbf{99.69}        & \textbf{99.29}      & \textbf{99.76}    & \textbf{99.10}         & \textbf{98.74}      \\
    \bottomrule
    \end{tabular}
    }
\label{tab:ap_GenVideo}
\end{table*}

\begin{table*}[ht]
    \centering
    \caption{Comparison of AUC (\%) between CMTA and baselines on GenVideo. \textsuperscript{\textdagger} Reproduced using the official code. \textsuperscript{\textdagger\textdagger} Reproduced from our implementation due to the unavailability of official code.}
    \resizebox{\textwidth}{!}{
    \begin{tabular}{c | c c c c c c c c c c | c}
    \toprule
    \rowcolor{gray!25}
    Method                                                         & MS     & MPS     & MV      & HotShot           & Show-1            & Gen2                 & Crafter                 & LaVie                 & Sora                 & WS                 & mean   \\
    \midrule 
    \hline
    \rowcolor{gray!5}
    STIL\textsuperscript{\textdagger} \cite{10.1145/3474085.3475508}, ACM MM 2021               & 86.48             & 88.75           & 82.77             & 60.00             & 68.71             & 86.67                & 73.56                   & 72.67                 & 54.18                & 69.94                      & 74.37  \\ 
    FTCN\textsuperscript{\textdagger} \cite{zheng2021exploring}, ICCV 2021                      & 69.76          & 83.25           & 97.18           & 88.69             & 93.45             & 93.33                & 91.62                   & 84.70               & 38.30                & 85.74                  & 82.60     \\
    \rowcolor{gray!5}
    X-CLIP\textsuperscript{\textdagger} \cite{ni2022expanding}, ECCV 2022                       & 80.51          & 86.14           & 95.12           & 91.51             & \underline{94.69}             & 88.96                & 92.84                   & 85.04               & 64.09                & \underline{87.93}                  & 88.87     \\  
    TALL\textsuperscript{\textdagger} \cite{xu2023tall}, ICCV 2023                              & 58.46             & 60.54           & 83.24             & 45.21             & 46.24             & 73.30                & 66.37                   & 48.40                 & 66.36                & 53.84                      & 60.20   \\ 
    \rowcolor{gray!5}  
    FID\textsuperscript{\textdagger} \cite{NEURIPS2024_6dddcff5}, NeurIPS 2024                  & 90.94             & 91.93           & 93.70             & 85.77 & 91.19 & 93.16                & 92.02                   & 82.57     & 73.45                & 81.30          & 87.60  \\ 
    NPR\textsuperscript{\textdagger} \cite{Tan_2024_CVPR}, CVPR 2024                            & 93.92 & \underline{99.38}  & \textbf{99.92}    & 26.83             & 18.99             & 98.78                & \underline{99.24}          & 42.32                 & \underline{97.56}    & 76.12                      & 75.31  \\
    \rowcolor{gray!5}
    MINTIME\textsuperscript{\textdagger} \cite{10547206}, TIFS 2024                            & 81.27             & 85.48              & 90.97              & 90.34              & 90.89             & 88.80              & 89.77                 & 83.86               & 79.27                & 85.71                  & 86.64               \\
    AIGVDet\textsuperscript{\textdagger} \cite{bai2024ai}, PRCV 2024                            & 68.05             & 79.42           & 59.41             & 74.67             & 70.29             & 71.62                & 69.77                   & 79.55                 & 60.79                & 67.82                      & 70.14  \\  
    \rowcolor{gray!5}
    DeMamba\textsuperscript{\textdagger} \cite{chen2024DeMamba}, arXiv 2024                     & 80.53             & 95.52           & \underline{99.90} & 59.49             & 49.34             & 99.10    & 97.99                   & 63.25                 & 91.76                & 71.15                      & 80.80  \\ 
    DeCoF\textsuperscript{\textdagger\textdagger} \cite{Ma2024DetectingAV}, ICME 2025   & 95.49          & 96.87           & 99.60           & 90.22             & 77.86             & 98.94                & 96.63                   & 84.62                 & 96.36                & 87.72                      & 92.43   \\
    \rowcolor{gray!5}
    NSG-VD\textsuperscript{\textdagger} \cite{zhang2025NSGVD}, NeurIPS 2025      & 69.76 & 83.25 & 97.18 & 88.69 & 93.45 & 93.33 & 91.62 & 84.70 & 38.30 & 85.74 & 82.60  \\
    ReStraV\textsuperscript{\textdagger} \cite{interno2025aigenerated}, NeurIPS 2025            & \underline{96.84}             & 98.20              & 96.95              & \textbf{99.27}              & 79.65             & \underline{99.93}             & 77.40             & \underline{97.75}             & 97.35             & 81.55             & \underline{92.49}              \\ 
    \rowcolor{gray!5}   
    D3\textsuperscript{\textdagger} \cite{Zheng_2025_ICCV}, ICCV 2025        & 86.00           & 93.62           & 95.74           & 96.96           & 94.03           & 93.93           & 96.21           & 88.93           & 88.93           & 87.60           & 92.20       \\  
    \hline \hline
    \rowcolor{gray!25}
    \textbf{CMTA}                                                                              & \textbf{99.41}    & \textbf{99.78}  & 97.92          & \underline{98.31}    & \textbf{98.64}    & \textbf{99.98}       & \textbf{99.51}       & \textbf{98.96}        & \textbf{99.74}       & \textbf{98.79}             & \textbf{99.10}  \\ 
    \bottomrule
    \end{tabular}
    }
\label{tab:auc_genvideo}
\end{table*}

\begin{table*}[th]
    \centering
    \caption{Comparison of AP ($\%$) between CMTA and baselines on EvalCrafter. \textsuperscript{\textdagger} Results reproduced using the official code. \textsuperscript{\textdagger\textdagger} Results reproduced from our implementation due to unavailable official code.}
    \resizebox{\textwidth}{!}{
    \begin{tabular}{c | c c c c c c c c c c c c c c| c}
    \toprule
    \rowcolor{gray!25}
    Method &  \makecell[c]{\rotatebox{50}{Floor33}} & \makecell[c]{\rotatebox{50}{Gen2}} &  \makecell[c]{\rotatebox{50}{Gen2-D}} &  \makecell[c]{\rotatebox{50}{HS-XL}} &  \makecell[c]{\rotatebox{50}{LaVie-B}} &  \makecell[c]{\rotatebox{50}{LaVie-I}} &  \makecell[c]{\rotatebox{50}{Mix-SR}} &  \makecell[c]{\rotatebox{50}{MS}} &  \makecell[c]{\rotatebox{50}{MV}} &  \makecell[c]{\rotatebox{50}{PKL}} &  \makecell[c]{\rotatebox{50}{PKL-V1}} &  \makecell[c]{\rotatebox{50}{Show-1}} &  \makecell[c]{\rotatebox{50}{VC}} & \makecell[c]{\rotatebox{50}{ZS}} &  \makecell[c]{\rotatebox{50}{mean}} \\
    \midrule
    \hline
    \rowcolor{gray!5}
    STIL\textsuperscript{\textdagger}               & 90.42        & 90.59        & 68.07        & 59.36        & 65.96        & 66.02        & 65.41        & 89.20        & 74.71        & 95.86        & 79.97        & 60.21        & 67.14        & 89.73        & 75.90        \\
    FTCN\textsuperscript{\textdagger}               & 82.85        & 88.50        & 94.72        & 88.25        & 82.71        & 81.91        & 92.74        & 71.16        & 96.65        & 96.36        & 93.14        & 93.55        & 92.23        & 69.09        & 87.42        \\
    \rowcolor{gray!5}
    X-CLIP\textsuperscript{\textdagger}             & 87.32        & 87.56        & 90.72        & 89.67        & 85.58        & 83.55        & 94.22        & 79.45        & 96.04        & 97.44        & 96.15        & 95.68        & 91.42        & 77.63        & 90.28        \\    
    TALL\textsuperscript{\textdagger}               & 63.25        & 70.75        & 77.04        & 46.93        & 52.87        & 52.53        & 78.16        & 62.11        & 83.63        & 65.33        & 70.98        & 48.00        & 60.50        & 51.73        & 63.13        \\
    \rowcolor{gray!5}  
    FID\textsuperscript{\textdagger}                & 96.40        & 97.36        & 98.68        & 89.90        & 92.92        & 84.19        & 98.51        & 95.74        & 98.29        & 99.49        & 99.17        & \underline{96.77}        & 95.71        & 95.18        & \underline{95.59}        \\ 
    NPR\textsuperscript{\textdagger}                & \textbf{99.77}        & 99.34        & \underline{99.95}        & 47.39        & 76.45        & 72.23        & \underline{99.67}   & \underline{98.54}        & \underline{99.96}        & \underline{99.97}        & \textbf{99.93}        & 69.82        & \underline{99.68}        & \underline{98.21}        & 90.07        \\
    \rowcolor{gray!5}
    MINTIME\textsuperscript{\textdagger}           & 84.62        & 86.20        & 88.02        & 90.07        & 83.12        & 82.99        & 88.87        & 78.47        & 90.56        & 94.65        & 91.31        & 87.99         & 88.18        & 88.43        & 87.39        \\  
    AIGVDet\textsuperscript{\textdagger}            & 67.84        & 71.86        & 74.24        & 51.46        & 73.81        & 70.72        & 57.64        & 71.00        & 56.50        & 94.95        & 92.92        & 72.41        & 64.58        & 67.00        & 70.50        \\ 
    \rowcolor{gray!5}  
    DeMamba\textsuperscript{\textdagger}            & 97.50        & 89.82        & 97.67        & 66.31        & 75.37        & 69.51        & 96.38        & 41.13        & 85.79        & 70.08        & 34.32        & 46.13        & 98.19        & 96.34        & 76.04        \\
    DeCoF\textsuperscript{\textdagger\textdagger}   & 91.69             & 96.30              & 97.18              & 76.02              & 70.40             & 59.39              & 93.77                 & 91.18               & 98.68                & 98.70                  & 98.43                 & 48.99               & 86.75                & 83.85                  & 85.10               \\
    \rowcolor{gray!5}
    NSG-VD\textsuperscript{\textdagger}             & 82.85 & 88.50 & 94.72 & 88.25 & 82.71 & 81.91 & 92.74 & 71.16 & 96.65 & 96.36 & 93.14 & 93.55 & 92.23 & 69.09 & 87.42  \\
    ReStraV\textsuperscript{\textdagger}            & 97.72             & \underline{99.93}              & 99.94              & \underline{99.21}              & \underline{95.70}             & \underline{97.35}             & 70.64             & 96.04             & 96.99             & \textbf{100.00}             & 99.71               & 77.14             & 75.18             & 63.45             & 90.64              \\ 
    \rowcolor{gray!5}  
    D3\textsuperscript{\textdagger}                 & 94.09           & 93.36           & 96.22           & 96.96           & 89.37           & 87.28           & 96.17           & 86.70           & 96.74           & 94.34           & 95.02           & 94.16           & 95.78           & 94.84           & 93.64       \\   
    \hline \hline
    \rowcolor{gray!25}
    \textbf{CMTA}                               & \underline{99.70}        & \textbf{99.98}        & \textbf{99.96}        & \textbf{99.41}        & \textbf{99.71}        & \textbf{99.50}        & \textbf{99.91}        & \textbf{99.60}        & \textbf{100.00}        & 99.93        & \underline{99.74}        & \textbf{99.88}        & \textbf{99.92}        & \textbf{98.98}        & \textbf{99.73}        \\  
    \bottomrule
    \end{tabular}
    }
\label{tab:ap_EvalCrafter}
\end{table*}

\begin{table*}[th]
    \centering
    \caption{Comparison of AUC (\%) between CMTA and baselines on EvalCrafter. \textsuperscript{\textdagger} Reproduced using the official code. \textsuperscript{\textdagger\textdagger} Reproduced from our implementation due to the unavailability of official code.}
    \resizebox{\textwidth}{!}{
    \begin{tabular}{c | c c c c c c c c c c c c c c| c}
    \toprule
    \rowcolor{gray!25}
    Method &  \makecell[c]{\rotatebox{50}{Floor33}} & \makecell[c]{\rotatebox{50}{Gen2}} &  \makecell[c]{\rotatebox{50}{Gen2-D}} &  \makecell[c]{\rotatebox{50}{HS-XL}} &  \makecell[c]{\rotatebox{50}{LaVie-B}} &  \makecell[c]{\rotatebox{50}{LaVie-I}} &  \makecell[c]{\rotatebox{50}{Mix-SR}} &  \makecell[c]{\rotatebox{50}{MS}} &  \makecell[c]{\rotatebox{50}{MV}} &  \makecell[c]{\rotatebox{50}{PKL}} &  \makecell[c]{\rotatebox{50}{PKL-V1}} &  \makecell[c]{\rotatebox{50}{Show-1}} &  \makecell[c]{\rotatebox{50}{VC}} & \makecell[c]{\rotatebox{50}{ZS}} &  \makecell[c]{\rotatebox{50}{mean}} \\
    \midrule 
    \hline
    \rowcolor{gray!5}
    STIL\textsuperscript{\textdagger}               & 90.14             & 91.40             & 79.25             & 61.62             & 73.63             & 71.23             & 74.22             & 87.28             & 84.22             & 97.82     & 88.50             & 66.68             & 74.49             & 90.77     & 80.80          \\  
    FTCN\textsuperscript{\textdagger}               & 82.78             & 90.76             & 95.11             & 90.29             & 84.51             & 83.00             & 92.31             & 70.99             & 96.65             & 96.73     & 94.01             & 93.62             & 91.27             & 70.82      & 88.06          \\
    \rowcolor{gray!5}
    X-CLIP\textsuperscript{\textdagger}             & 86.10             & 86.70             & 90.93             & 90.80             & 84.06             & 83.01             & 93.59             & 79.69             & 95.56             & 97.45     & 95.79             & 95.29             & 90.71             & 80.53      & 89.38          \\  
    TALL\textsuperscript{\textdagger}               & 61.46             & 72.04             & 74.60             & 43.88             & 49.64             & 50.02             & 77.50             & 58.99             & 83.19             & 67.13             & 72.51             & 46.81             & 55.69             & 46.03             & 61.39             \\  
    \rowcolor{gray!5} 
    FID\textsuperscript{\textdagger}                & 96.26             & 97.46             & 98.58             & 90.05             & 92.32             & 83.42             & 98.23             & 95.27     & 98.28 & 99.48     & 99.24 & \underline{96.79} & 95.26 & \underline{95.29}     & \underline{95.42}     \\ 
    NPR\textsuperscript{\textdagger}                & \underline{99.38} & 97.74             & 99.80             & 26.83             & 50.23             & 36.00             & \underline{99.22}             & 93.92             & \underline{99.92}     & \underline{99.92}     & \underline{99.71}     & 18.99             & \underline{99.25}     & 92.60     & 79.54             \\
    \rowcolor{gray!5}
    MINTIME\textsuperscript{\textdagger}           & 86.88             & 88.01             & 89.41             & 91.71             & 84.07             & 84.23             & 90.16             & 79.60             & 91.68             & 95.08      & 92.33            & 90.08               & 90.16            & 91.29      & 88.91               \\    
    AIGVDet\textsuperscript{\textdagger}            & 79.52             & 70.10             & 73.03             & 73.92             & 85.34 & 74.73 & 64.44             & 68.17             & 59.90             & 92.78             & 90.60             & 70.23             & 74.92             & 67.75             & 74.67             \\  
    \rowcolor{gray!5} 
    DeMamba\textsuperscript{\textdagger}                      & 95.41             & 98.60     & 99.70     & 59.10             & 72.87             & 55.39             & 98.91     & 79.11             & 99.87     & 99.51 & 99.70     & 48.89             & 97.08             & 76.98             & 84.37             \\ 
    DeCoF\textsuperscript{\textdagger\textdagger}   & 96.87             & 98.69              & 99.17              & 90.22              & 87.71             & 81.53              & 98.05                 & 95.49               & 99.60                & 99.54                  & 99.42                 & 77.89               & 95.20                & 92.20                  & 93.68               \\
    \rowcolor{gray!5}
    NSG-VD\textsuperscript{\textdagger}             & 82.78 & 90.76 & 95.11 & 90.29 & 84.51 & 83.00 & 92.31 & 70.99 & 96.65 & 96.73 & 94.01 & 93.62 & 91.27 & 70.82 & 88.06  \\
    ReStraV\textsuperscript{\textdagger}            & 98.33             & \underline{99.93}              & \underline{99.94}              & \underline{99.13}              & \underline{96.93}             & \underline{97.97}              & 73.98              & \underline{97.43}              & 96.53             & \textbf{100.00}              & \textbf{99.85}              & 81.71              & 82.05             & 68.22             & 92.29              \\ 
    \rowcolor{gray!5}
    D3\textsuperscript{\textdagger}                 & 93.68           & 92.24           & 95.85           & 96.93           & 90.06           & 88.68           & 95.62           & 87.38           & 96.32           & 93.77           & 94.73           & 92.49           & 95.58           & 93.76           & 93.36              \\  
    \hline \hline
    \rowcolor{gray!25}
    \textbf{CMTA}                               & \textbf{99.63}  & \textbf{99.98}     & \textbf{99.96} & \textbf{99.60}     & \textbf{99.61} & \textbf{99.31} & \textbf{99.89}     & \textbf{99.47} & \textbf{100.00} & 99.90             & 99.64             & \textbf{99.85}     & \textbf{99.90}     & \textbf{99.47}             & \textbf{99.73}     \\  
    \bottomrule
    \end{tabular}
    }
\label{tab:auc_EvalCrafter}
\end{table*}

\begin{table*}[th]
    \centering
    \caption{Comparison of AP ($\%$) between CMTA and baselines on VideoPhy.  \textsuperscript{\textdagger} Results reproduced using the official code. \textsuperscript{\textdagger\textdagger} Results reproduced from our implementation due to unavailable official code.}
    \resizebox{\textwidth}{!}{
    \begin{tabular}{c | c c c c c c c c c c| c}
    \toprule
    \rowcolor{gray!25}
    Method                                                                                                 & CVX   & CVX-5B   & DM   & Gen-2        & LaVie        & OpenSora     & Pika       & SVD-T2I2V    & VC2     & ZS    & mean   \\
    \midrule
    \hline
    \rowcolor{gray!5}
    STIL\textsuperscript{\textdagger} \cite{10.1145/3474085.3475508}, ACM MM 2021                          & 66.43        & 69.61         & 65.31           & 69.63        & 63.68        & 61.94        & 97.07      & 67.51        & 67.54             & 58.56        & 68.73   \\
    FTCN\textsuperscript{\textdagger} \cite{zheng2021exploring}, ICCV 2021                                 & 74.24        & 74.83         & 67.98           & 93.99        & 69.16        & 66.95        & 93.94      & 82.81        & 83.39             & 75.64        & 78.29   \\
    \rowcolor{gray!5}
    X-CLIP\textsuperscript{\textdagger} \cite{ni2022expanding}, ECCV 2022                                  & 87.35        & 83.72         & 75.54           & 92.09        & 79.52        & 90.44        & 96.13      & 85.26        & 85.35             & 88.88        & 86.66   \\ 
    TALL\textsuperscript{\textdagger} \cite{xu2023tall}, ICCV 2023                                         & 39.59        & 50.72         & 62.36           & 70.78        & 40.40        & 37.30        & 62.69      & 52.62        & 52.66             & 50.66        & 51.98   \\
    \rowcolor{gray!5}
    FID\textsuperscript{\textdagger} \cite{NEURIPS2024_6dddcff5}, NeurIPS 2024                             & \underline{93.34} & \underline{91.41}         & 97.50   & 98.35 & 96.51 & 87.90 & 99.55 & 95.66 & 96.03     & 90.60 & \underline{94.69} \\
    NPR\textsuperscript{\textdagger} \cite{Tan_2024_CVPR}, CVPR 2024                                       & 81.37        & 81.99         & \underline{99.86}   & \underline{99.90} & 63.72        & 88.78 & \underline{99.91} & \textbf{99.54}        & 60.21             & 78.23        & 85.35   \\
    \rowcolor{gray!5}
    MINTIME\textsuperscript{\textdagger} \cite{10547206}, TIFS 2024                                       & 85.27             & 85.96              & 77.60              & 84.12              & 79.16             & 90.64              & 92.86                 & 79.01               & 83.83                & 86.34                  & 84.48               \\ 
    AIGVDet\textsuperscript{\textdagger} \cite{bai2024ai}, PRCV 2024                                       & 63.15        & 58.95         & 59.27           & 61.55        & 61.06        & 59.07        & 92.96      & 53.73        & 58.22             & 63.11        & 63.11   \\
    \rowcolor{gray!5}
    DeMamba\textsuperscript{\textdagger} \cite{chen2024DeMamba}, arXiv 2024                                & 22.10        & 15.96         & 72.52           & 92.92        & 50.00        & \underline{94.70}        & 45.47      & 91.59        & 90.15             & 73.64        & 64.91   \\
    DeCoF\textsuperscript{\textdagger\textdagger} \cite{Ma2024DetectingAV}, ICME 2025   & 14.35             & 17.10              & 79.20              & 88.81              & 42.72             & 19.92              & 95.01                 & 59.68               & 46.13                & 21.59                  & 48.45               \\
    \rowcolor{gray!5}
    NSG-VD\textsuperscript{\textdagger} \cite{zhang2025NSGVD}, NeurIPS 2025               & 74.24 & 74.83 & 67.98 & 93.99 & 69.16 & 66.95 & 93.94 & 82.81 & 83.39 & 75.64 & 78.29  \\
    ReStraV\textsuperscript{\textdagger} \cite{interno2025aigenerated}, NeurIPS 2025            & 72.57             & 61.39              & 99.09              & \textbf{99.95}              & \textbf{98.61}             & \textbf{99.57}             & \textbf{100.00}             & 45.01             & \textbf{99.42}             & \textbf{98.55}             & 87.42              \\ 
    \rowcolor{gray!5}
    D3\textsuperscript{\textdagger} \cite{Zheng_2025_ICCV}, ICCV 2025                                      & 90.23        & \textbf{93.94}         & 96.25           & 94.88           & 82.22           & 92.38           & 93.21           & 95.60           & 92.50           & 92.52           & 92.37              \\   
    \hline \hline
    \rowcolor{gray!25}
    \textbf{CMTA}                                                                                      & \textbf{94.28} & 89.64 & \textbf{100.00}   & 97.19        & \underline{97.84} & 91.24        & 93.97      & \underline{97.75} & \underline{98.66}     & \underline{94.82} & \textbf{95.54} \\ 
    \bottomrule
    \end{tabular}
    }
\label{tab:ap_VideoPhy}
\end{table*}

\begin{table*}[th]
    \centering
    \caption{Comparison of AUC ($\%$) between CMTA and baselines on VideoPhy.  \textsuperscript{\textdagger} Results reproduced using the official code. \textsuperscript{\textdagger\textdagger} Results reproduced from our implementation due to unavailable official code.}
    \resizebox{\textwidth}{!}{
    \begin{tabular}{c | c c c c c c c c c c | c}
    \toprule
    \rowcolor{gray!25}
    Method                                                                                                 & CVX    & CVX-5B & DM   & Gen-2   & LaVie   & OpenSora      & Pika    & SVD-T2I2V   & VC2   & ZS    & mean   \\
    \midrule
    \hline 
    \rowcolor{gray!5}
    STIL\textsuperscript{\textdagger} \cite{10.1145/3474085.3475508}, ACM MM 2021                          & 71.73        & 73.94        & 73.01           & 80.75   & 68.66   & 63.42        & 98.47 & 77.82       & 75.18           & 55.85        & 73.88        \\
    FTCN\textsuperscript{\textdagger} \cite{zheng2021exploring}, ICCV 2021                                 & 80.91        & 81.01        & 70.20           & 95.21   & 72.88   & 76.37        & 95.90 & 87.07       & 86.53           & 74.88        & 82.10        \\
    \rowcolor{gray!5}
    X-CLIP\textsuperscript{\textdagger} \cite{ni2022expanding}, ECCV 2022                                  & 88.99        & 84.84        & 78.57           & 93.35   & 76.51   & 92.68        & 96.22 & 84.92       & 86.27           & 88.06        & 86.88        \\ 
    TALL\textsuperscript{\textdagger} \cite{xu2023tall}, ICCV 2023                                         & 30.05        & 46.36        & 56.54           & 64.80   & 32.94   & 24.82        & 63.65    & 45.40       & 42.48           & 39.44        & 44.65        \\ 
    \rowcolor{gray!5}
    FID\textsuperscript{\textdagger} \cite{NEURIPS2024_6dddcff5}, NeurIPS 2024                             & \underline{93.37} & 91.72        & 97.62   & 98.11 & 96.03 & 86.03 & 99.59 & 95.81 & 95.68     & 89.38 & \underline{94.33} \\
    NPR\textsuperscript{\textdagger} \cite{Tan_2024_CVPR}, CVPR 2024                                       & 72.10        & 73.60        & \underline{99.70}   & \underline{99.80} & 42.90   & 83.50        & \underline{99.80} & \textbf{99.50}       & 47.20           & 52.90        & 77.10        \\
    \rowcolor{gray!5}
    MINTIME\textsuperscript{\textdagger} \cite{10547206}, TIFS 2024                                       & 85.08        & 87.24        & 82.06           & 86.32          & 76.29   & 89.92        & 93.61          & 80.36                & 85.09           & 89.30        & 85.53        \\  
    AIGVDet\textsuperscript{\textdagger} \cite{bai2024ai}, PRCV 2024                                       & 78.13        & 67.05        & 68.40           & 75.87   & 71.25   & 69.57        & 97.83    & 72.87       & 81.36           & 77.15        & 75.95        \\
    \rowcolor{gray!5}
    DeMamba\textsuperscript{\textdagger} \cite{chen2024DeMamba}, arXiv 2024                                          & 59.65        & 61.44        & 98.42           & 99.64   & 63.66   & 51.32        & 99.19    & 97.72 & 63.57           & 46.56        & 74.12        \\ 
    DeCoF\textsuperscript{\textdagger\textdagger} \cite{Ma2024DetectingAV}, ICME 2025   & 53.29             & 60.92              & 96.21              & 98.64              & 82.43             & 64.90              & 99.10                 & 92.99               & 84.98                & 63.71                  & 79.72               \\
    \rowcolor{gray!5}
    NSG-VD\textsuperscript{\textdagger} \cite{zhang2025NSGVD}, NeurIPS 2025            & 80.91 & 81.01 & 70.20 & 95.21 & 72.88 & 76.37 & 95.90 & 87.07 & 86.53 & 74.88 & 82.10  \\
    ReStraV\textsuperscript{\textdagger} \cite{interno2025aigenerated}, NeurIPS 2025            & 78.09             & 65.21              & 98.83              & \textbf{99.94}              & \textbf{98.91}             & \textbf{99.65}             & \textbf{100.00}             & 41.02             & \textbf{99.44}             & \textbf{99.01}             & 88.01              \\ 
    \rowcolor{gray!5} 
    D3\textsuperscript{\textdagger} \cite{Zheng_2025_ICCV}, ICCV 2025                                      & 90.05           & \textbf{93.51}           & 95.56           & 94.34           & 84.26           & 91.90           & 92.02           & 95.09           & 91.68           & 90.91           & 91.93            \\  
    \hline \hline
    \rowcolor{gray!25}
    \textbf{CMTA}                                                                                      & \textbf{96.93} & \underline{93.23} & \textbf{100.00}           & 98.51   & \underline{98.65} & \underline{94.51} & 96.82 & \underline{98.57}       & \underline{99.31}     & \underline{96.92} & \textbf{97.34} \\ 
    \bottomrule
    \end{tabular}
    }
\label{tab:auc_VideoPhy}
\end{table*}

\begin{table*}[th]
    \centering
    \caption{Comparison of AP ($\%$) between CMTA and baselines on VidProM. \textsuperscript{\textdagger} Results reproduced using the official code. \textsuperscript{\textdagger\textdagger} Results reproduced from our implementation due to unavailable official code.}
    \resizebox{0.70\textwidth}{!}{
    \begin{tabular}{c | c c c c c c | c}
    \toprule
    \rowcolor{gray!25}
    Method                                                                                    & MS   & OpenSora   & Pika        & ST2V     &  T2VZ    & VC2     & mean \\
    \midrule
    \hline
    \rowcolor{gray!5}
    STIL\textsuperscript{\textdagger} \cite{10.1145/3474085.3475508}, ACM MM 2021             & 43.68       & 63.15       & 95.28 & 55.46       & 48.68       & 61.49       & 61.29       \\
    FTCN\textsuperscript{\textdagger} \cite{zheng2021exploring}, ICCV 2021                    & 74.65       & 90.05       & 94.26       & 83.53       & 46.25       & 87.27       & 79.33       \\
    \rowcolor{gray!5}
    X-CLIP\textsuperscript{\textdagger} \cite{ni2022expanding}, ECCV 2022                     & 82.24       & \underline{93.47}       & 95.80       & 86.42       & 70.92       & 91.60       & 88.05       \\ 
    TALL\textsuperscript{\textdagger} \cite{xu2023tall}, ICCV 2023                            & 50.93       & 54.50       & 63.47       & 51.50       & 60.99       & 59.70       & 56.85       \\
    \rowcolor{gray!5}
    FID\textsuperscript{\textdagger} \cite{NEURIPS2024_6dddcff5}, NeurIPS 2024                & 91.35 & 87.68       & 99.59 & \textbf{97.87} & 68.51       & 85.92 & 88.49 \\
    NPR\textsuperscript{\textdagger} \cite{Tan_2024_CVPR}, CVPR 2024                          & 87.04       & 89.85 & \textbf{99.98} & 89.88 & 88.93 & 70.79       & 87.75       \\
    \rowcolor{gray!5}
    MINTIME\textsuperscript{\textdagger} \cite{10547206}, TIFS 2024                          & 76.09       & 88.32       & 94.19       & 50.03       & 65.06       & 86.77       & 76.74       \\ 
    AIGVDet\textsuperscript{\textdagger} \cite{bai2024ai}, PRCV 2024                          & 63.33       & 62.12       & 66.07       & 55.46       & 63.49       & 52.15       & 60.44       \\
    \rowcolor{gray!5}
    DeMamba\textsuperscript{\textdagger} \cite{chen2024DeMamba}, arXiv 2024                   & \textbf{99.24}       & 48.00       & 86.95       & 33.94       & \textbf{98.47}       & \textbf{98.38}       & 77.50       \\
    DeCoF\textsuperscript{\textdagger\textdagger} \cite{Ma2024DetectingAV}, ICME 2025         & 87.93       & 85.42       & 99.56       & 76.27       & 93.72       & 96.75       & \underline{90.11}       \\
    \rowcolor{gray!5}
    NSG-VD\textsuperscript{\textdagger} \cite{zhang2025NSGVD}, NeurIPS 2025                   & 74.65 & 90.05 & 94.26 & 83.53 & 46.25 & 87.27 & 79.33  \\
    ReStraV\textsuperscript{\textdagger} \cite{interno2025aigenerated}, NeurIPS 2025          & \underline{96.99}             & \textbf{97.20}             & \underline{99.85}             & 63.96             & 39.63             & 69.14             & 77.80              \\ 
    \rowcolor{gray!5}
    D3\textsuperscript{\textdagger} \cite{Zheng_2025_ICCV}, ICCV 2025                         & 87.61       & 90.73       & 93.17       & 81.00       & 61.93       & 93.97       & 84.73         \\  
    \hline \hline
    \rowcolor{gray!25}
    \textbf{CMTA}                                                                             & 86.19 & 91.06 & 97.58 & \underline{95.94}       & \underline{98.14} & \underline{97.17} & \textbf{94.35} \\ 
    \bottomrule
    \end{tabular}
    }
\label{tab:ap_VidProM}
\end{table*}

\begin{table*}[th]
    \centering
    \caption{Comparison of AUC ($\%$) between CMTA and baselines on VidProM. \textsuperscript{\textdagger} Results reproduced using the official code. \textsuperscript{\textdagger\textdagger} Results reproduced from our implementation due to unavailable official code.}
    \resizebox{0.7\textwidth}{!}{
    \begin{tabular}{c | c c c c c c | c}
    \toprule
    \rowcolor{gray!25}
    Method                                                                                                 & MS   & OpenSora   & Pika        & ST2V     &  T2VZ    & VC2     & mean \\
    \midrule 
    \hline
    \rowcolor{gray!5}
    STIL\textsuperscript{\textdagger} \cite{10.1145/3474085.3475508}, ACM MM 2021                          & 41.05       & 72.01       & 97.42 & 67.12       & 47.66       & 67.98       & 65.54       \\
    FTCN\textsuperscript{\textdagger} \cite{zheng2021exploring}, ICCV 2021                                 & 70.98        & 90.76       & 95.23       & 87.81       & 50.70       & 88.43       & 80.65       \\
    \rowcolor{gray!5}
    X-CLIP\textsuperscript{\textdagger} \cite{ni2022expanding}, ECCV 2022                                  & 80.59        & 93.02       & 95.85       & 87.47       & 75.86       & 91.24       & 87.34       \\ 
    TALL\textsuperscript{\textdagger} \cite{xu2023tall}, ICCV 2023                                         & 45.29       & 56.39       & 66.91       & 50.31       & 57.80       & 53.48       & 55.03       \\ 
    \rowcolor{gray!5}
    FID\textsuperscript{\textdagger} \cite{NEURIPS2024_6dddcff5}, NeurIPS 2024                             & \underline{89.86} & 87.36       & 99.62 & \underline{98.10} & 67.01       & 85.65 & 87.93 \\
    NPR\textsuperscript{\textdagger} \cite{Tan_2024_CVPR}, CVPR 2024                                       & 82.61       & \textbf{98.56} & \underline{99.84} & \textbf{98.92} & 93.32 & 56.70       & 88.33 \\ 
    \rowcolor{gray!5}
    MINTIME\textsuperscript{\textdagger} \cite{10547206}, TIFS 2024                                       & 79.40        & 88.99       & 94.81       & 54.21       & 70.92       & 88.96       & 79.55       \\ 
    AIGVDet\textsuperscript{\textdagger} \cite{bai2024ai}, PRCV 2024                                       & 60.11       & 45.27       & 48.75       & 34.14       & 59.96       & 46.21       & 49.07       \\
    \rowcolor{gray!5}
    DeMamba\textsuperscript{\textdagger} \cite{chen2024DeMamba}, arXiv 2024                                & 54.44       & 84.02       & 99.29       & 84.94       & 76.34       & 78.61       & 79.61       \\ 
    DeCoF\textsuperscript{\textdagger\textdagger} \cite{Ma2024DetectingAV}, ICME 2025                      & 89.18        & 86.98       & 99.56       & 80.14       & \underline{93.85}       & \underline{96.94}       & \underline{90.77}       \\
    \rowcolor{gray!5}
    NSG-VD\textsuperscript{\textdagger} \cite{zhang2025NSGVD}, NeurIPS 2025            & 70.98 & 90.76 & 95.23 & 87.81 & 50.70 & 88.43 & 80.65  \\
    ReStraV\textsuperscript{\textdagger} \cite{interno2025aigenerated}, NeurIPS 2025                       & \textbf{97.96}             & \underline{97.53}             & \textbf{99.87}             & 68.44             & 28.60             & 73.79             & 77.70              \\  
    \rowcolor{gray!5}
    D3\textsuperscript{\textdagger} \cite{Zheng_2025_ICCV}, ICCV 2025                                      & 87.21        & 91.29       & 92.24       & 79.74       & 56.31       & 92.77       & 83.26         \\  
    \hline \hline
    \rowcolor{gray!25}
    \textbf{CMTA}                                                                                          & 88.15 & 94.16 & 98.71 & 97.14       & \textbf{98.79} & \textbf{98.19} & \textbf{95.86} \\ 
    \bottomrule
    \end{tabular}
    }
\label{tab:auc_VidProM}
\end{table*}

\begin{table}[!ht]
    \centering
    \caption{Comparison of ACC (\%) of CMTA and baselines on GenVideo, EvalCrafter, VideoPhy, and VidProM.
    \textsuperscript{\textdagger} Results reproduced using the official code. \textsuperscript{\textdagger\textdagger} Results reproduced from our implementation due to unavailable official code.}
    \resizebox{\columnwidth}{!}{
    \begin{tabular}{c | c c c c }
    \toprule
    \rowcolor{gray!25}
    Method                                                                                                 & GenVideo          & EvalCrafter     & VideoPhy         & VidProM    \\
    \midrule
    \hline
    \rowcolor{gray!5}
    STIL\textsuperscript{\textdagger} \cite{10.1145/3474085.3475508}, ACM MM 2021                          & 59.90             & 66.78           & 56.62            & 56.58      \\
    FTCN\textsuperscript{\textdagger} \cite{zheng2021exploring}, ICCV 2021                                 & 70.52           & 73.87           & 62.43            & 67.23      \\
    \rowcolor{gray!5}
    X-CLIP\textsuperscript{\textdagger} \cite{ni2022expanding}, ECCV 2022                                  & 75.34           & 76.82           & 69.26            & 73.37      \\   
    TALL\textsuperscript{\textdagger} \cite{xu2023tall}, ICCV 2023                                         & 57.47             & 58.42           & 48.78            & 54.02      \\ 
    \rowcolor{gray!5} 
    FID\textsuperscript{\textdagger} \cite{NEURIPS2024_6dddcff5}, NeurIPS 2024                             & 54.57             & 63.59           & 65.01            & 54.44      \\  
    NPR\textsuperscript{\textdagger} \cite{Tan_2024_CVPR}, CVPR 2024                                       & 65.41             & 71.36           & 57.00            & 68.04      \\ 
    \rowcolor{gray!5}
    MINTIME\textsuperscript{\textdagger} \cite{10547206}, TIFS 2024                                       & 78.55           & 81.52           & \underline{77.13}            & 71.74      \\  
    AIGVDet\textsuperscript{\textdagger} \cite{bai2024ai}, PRCV 2024                                       & 49.07             & 57.62           & 53.33            & 47.25      \\  
    \rowcolor{gray!5}
    DeMamba\textsuperscript{\textdagger} \cite{chen2024DeMamba}, arXiv 2024                                & 54.12             & 62.45           & 42.29            & 42.59      \\  
    DeCoF\textsuperscript{\textdagger\textdagger} \cite{Ma2024DetectingAV}, ICME 2025                      & \underline{87.60}           & \underline{89.75}  & 63.71  & \underline{85.47}      \\
    \rowcolor{gray!5}
    NSG-VD\textsuperscript{\textdagger} \cite{zhang2025NSGVD}, NeurIPS 2025                                & 70.52           & 73.87           & 62.43            & 67.23      \\
    ReStraV\textsuperscript{\textdagger} \cite{interno2025aigenerated}, NeurIPS 2025                       & 56.17             & 63.93              & 62.00              & 57.13              \\  
    \rowcolor{gray!5} 
    D3\textsuperscript{\textdagger} \cite{Zheng_2025_ICCV}, ICCV 2025                                      & 76.19           & 76.88           & 72.84            & 64.16      \\      
    \hline \hline
    \rowcolor{gray!25}
    \textbf{CMTA}                                                                                          & \textbf{97.39}    & \textbf{98.26}       & \textbf{93.66}        & \textbf{89.35}  \\   
    \bottomrule
    \end{tabular}
    }
\label{tab:acc_all}
\end{table}

\begin{table*}[!ht]
    \centering
    \caption{Ablation performance of individual components in CMTA on GenVideo, reported in terms of AP (\%).}
    \resizebox{\textwidth}{!}{
    \begin{tabular}{c c c c| c c c c c c c c c c | c}
    \toprule
    Visual          & Textual         & CGTM     & FGTM         & MS     & MPS     & MV     & HotShot     & Show-1     & Gen2     & Crafter     & LaVie     & Sora     & WS     & mean   \\
    \midrule
    \hline
    $\checkmark$    & $\times$        & $\times$       & $\times$            & \underline{95.12}          & \underline{96.32}           & \textbf{97.88}          & \underline{89.35}       & \underline{91.40}      & \underline{95.95}    & \underline{95.03}       & \underline{86.23}     & \underline{99.75}    & \underline{89.89}          & \underline{93.69}  \\ 
    $\times$        & $\checkmark$    & $\times$       & $\times$            & 65.27          & 67.29           & 66.72          & 64.14       & 64.47      & 64.78    & 66.49       & 64.32     & 90.25    & 66.99          & 68.07  \\ 
    $\checkmark$    & $\checkmark$    & $\checkmark$   & $\times$            & 69.04          & 68.88           & 84.93          & 76.25       & 79.79      & 80.77    & 80.56       & 76.31     & 79.89    & 66.63          & 76.31  \\ 
    $\checkmark$    & $\checkmark$    & $\times$       & $\checkmark$        & 92.36          & 93.56           & 94.15          & 85.19       & 87.84      & 92.23    & 91.02       & 79.78     & 98.79    & 83.32          & 90.82  \\ 
    \hline \hline
    \rowcolor{gray!25}
    $\checkmark$    & $\checkmark$    & $\checkmark$   & $\checkmark$        & \textbf{99.22}          & \textbf{99.59}           & \underline{96.09}          & \textbf{96.99}       & \textbf{97.65}      & \textbf{99.98}    & \textbf{99.69}       & \textbf{99.29}     & \textbf{99.76}    & \textbf{99.10}          & \textbf{98.74}  \\ 
    \bottomrule
    \end{tabular}
    }
\label{tab:ablation_ap}
\end{table*}

\begin{table*}[ht]
    \centering
    \caption{Ablation performance of individual components in CMTA on GenVideo, reported in terms of AUC (\%).}
    \resizebox{\textwidth}{!}{
    \begin{tabular}{c c c c| c c c c c c c c c c | c}
    \toprule
    Visual          & Textual         & CGTM     & FGTM         & MS     & MPS     & MV      & HotShot     & Show-1     & Gen2     & Crafter     & LaVie     & Sora     & WS     & mean   \\
    \midrule
    \hline
    $\checkmark$    & $\times$        & $\times$       & $\times$            & \underline{95.71}          & \underline{96.90}           & \textbf{98.60}           & \underline{90.15}       & \underline{92.30}      & \underline{96.97}    & \underline{96.00}       & \underline{86.27}     & \textbf{99.74}    & \underline{88.65}          & \underline{94.13}  \\ 
    $\times$        & $\checkmark$    & $\times$       & $\times$            & 66.10          & 68.19           & 66.28           & 64.69       & 64.87      & 64.58    & 64.19       & 62.57     & 90.56    & 63.95          & 67.70  \\ 
    $\checkmark$    & $\checkmark$    & $\checkmark$   & $\times$            & 70.41          & 69.78           & 82.97           & 76.57       & 80.07      & 80.87    & 80.65       & 76.90     & 80.74    & 67.48          & 77.64  \\ 
    $\checkmark$    & $\checkmark$    & $\times$       & $\checkmark$        & 92.66          & 93.90           & 95.68           & 83.83       & 87.92      & 93.31    & 92.14       & 77.65     & \underline{98.72}    & 79.67          & 90.75  \\ 
    \hline \hline
    \rowcolor{gray!25}
    $\checkmark$    & $\checkmark$    & $\checkmark$   & $\checkmark$        & \textbf{99.41} & \textbf{99.78}  & \underline{97.92}  & \textbf{98.31}  & \textbf{98.64}    & \textbf{99.98}    & \textbf{99.51} & \textbf{98.96} & \textbf{99.74}    & \textbf{98.79}          & \textbf{99.10}  \\ 
    \bottomrule
    \end{tabular}
    }
\label{tab:ablation_auc}
\end{table*}

\begin{table*}[ht]
    \centering
    \caption{Ablation performance of individual components in CMTA on GenVideo, reported in terms of ACC (\%).}
    \resizebox{\textwidth}{!}{
    \begin{tabular}{c c c c| c c c c c c c c c c | c}
    \toprule
    Visual          & Textual         & CGTM     & FGTM         & MS     & MPS     & MV     & HotShot     & Show-1     & Gen2     & Crafter     & LaVie     & Sora     & WS     & mean   \\
    \midrule
    \hline
    $\checkmark$    & $\times$        & $\times$       & $\times$            & \underline{85.14}          & \underline{89.57}           & \textbf{94.41}          & \underline{75.14}       & \underline{77.43}      & \underline{92.32}    & \underline{88.98}       & \underline{73.79}     & \textbf{96.43}    & \underline{78.01}          & \underline{85.62}  \\ 
    $\times$        & $\checkmark$    & $\times$       & $\times$            & 60.86          & 61.86           & 59.74          & 60.14       & 58.71      & 59.78    & 57.08       & 56.64     & 78.57    & 57.37          & 61.08  \\ 
    $\checkmark$    & $\checkmark$    & $\checkmark$   & $\times$            & 66.86          & 62.71           & 76.04          & 68.57       & 73.57      & 72.97    & 74.32       & 71.64     & 71.43    & 62.61          & 70.07  \\ 
    $\checkmark$    & $\checkmark$    & $\times$       & $\checkmark$        & 83.86          & 85.00           & 88.50          & 72.43       & 75.57      & 85.94    & 83.69       & 71.43     & \underline{92.86}    & 73.88          & 81.32  \\ 
    \hline \hline
    \rowcolor{gray!25}
    $\checkmark$    & $\checkmark$    & $\checkmark$   & $\checkmark$        & \textbf{98.29} & \textbf{99.71}  & \underline{94.09} & \textbf{97.86}  & \textbf{98.14}  & \textbf{99.49}  & \textbf{98.93}   & \textbf{97.14}   & \textbf{96.43}    & \textbf{93.86}     & \textbf{97.39}  \\ 
    \bottomrule
    \end{tabular}
    }
\label{tab:ablation_acc}
\end{table*}

\section{Experiments}
\label{sec:exper}

\subsection{Datasets}
To comprehensively evaluate the effectiveness of the proposed CMTA method, we conduct extensive experiments across four large-scale datasets: GenVideo~\cite{chen2024DeMamba}, EvalCrafter~\cite{liu2024evalcrafter}, VideoPhy~\cite{bansal2024videophy}, and VidProM~\cite{wang2024vidprom}, covering diverse video generation models and real-world scenarios to assess accuracy and generalization.

\noindent\textbf{Training Set.}
The training data comprise real videos from Youku-mPLUG~\cite{xu2023youku} and AI-generated videos produced by the Pika~\cite{pika2022} model, both of which are included in the GenVideo~\cite{chen2024DeMamba} training partition. 

\noindent\textbf{Validation Set.}
We randomly select 10\% of the training data as the validation set, which is used exclusively for adaptive learning-rate scheduling and hyperparameter tuning during training.

\noindent\textbf{Testing Set.}
The testing set includes 40 test subsets from four large-scale out-of-distribution benchmarks: GenVideo, EvalCrafter, VideoPhy, and VidProM, synthesized by various contemporary video generators.

\begin{itemize}
    \item \textbf{GenVideo}~\cite{chen2024DeMamba}: contains 10 subsets generated by ModelScope (MS) ~\cite{wang2023modelscope}, MorphStudio (MPS) ~\cite{morph2023}, MoonValley (MV) ~\cite{moonvalley2022}, HotShot~\cite{hotshot2023}, Show-1~\cite{zhang2024show}, Gen2~\cite{esser2023structure}, Crafter~\cite{chen2023videocrafter1}, LaVie~\cite{wang2025LaVie}, Sora~\cite{brooks2024video}, and WildScrape (WS) ~\cite{wei2024dreamvideo, feng2023dreamoving, xu2024magicanimate}.

    \item \textbf{EvalCrafter}~\cite{liu2024evalcrafter}: consists of 14 subsets, including MoonValley (MV), VideoCrafter V0.9 (Floor33), Gen2, Gen2-December (Gen2-D) , HotShot-XL (HS-XL), LaVie-Base (LaVie-B), LaVie-Interpolation (LaVie-I), Mix-SR, ModelScope (MS), PikaLab (PKL), PikaLab V1.0 (PKL-V1), Show-1, VideoCrafter (VC), and ZeroScope (ZS).

    \item \textbf{VideoPhy}~\cite{bansal2024videophy}: comprises 10 subsets generated by CogVideoX (CVX), CogVideoX-5B (CVX-5B), Dream-Machine (DM), Gen2, LaVie, OpenSora, Pika, SVD-T2I2V, VideoCrafter2 (VC2), and ZeroScope (ZS).

    \item \textbf{VidProM}~\cite{wang2024vidprom}: includes 6 subsets generated by ModelScope (MS), OpenSora, Pika, StreamingT2V (ST2V), Text2Video-Zero (T2VZ), and VideoCrafter2 (VC2).
\end{itemize}

Across all testing datasets, real video samples are uniformly drawn from MSR-VTT~\cite{xu2016msr}, following the same protocol established in the GenVideo benchmark~\cite{chen2024DeMamba}.

\subsection{Implementation Details}
We train CMTA for 200 epochs on a single NVIDIA RTX 4090 GPU, using a 1:1 balanced dataset of AI-generated and real video samples.
For each video, we randomly sample one contiguous clip of 8 frames. For each frame, we generate captions using the BLIP model with the pre-trained ``blip-image-captioning-base'' checkpoint, and then extract visual and textual embeddings using the frozen CLIP encoder with the pre-trained ``clip-vit-base-patch32'' checkpoint. All these pre-trained components are kept frozen during both training and testing. 
The obtained visual-textual features are fed into a two-layer Transformer encoder with four attention heads, which is trained from scratch with weights initialized by the Xavier uniform distribution and biases fixed to zero. 
The hidden dimension of the Transformer encoder is set to 256.
For coarse-grained temporal modeling, we adopt a GRU network with a 256-dimensional hidden state.
The entire model is optimized by the Adam optimizer with an initial learning rate of $1\times10^{-4}$ and a batch size of 256. 
The learning rate is dynamically adjusted according to validation performance using a ``ReduceLROnPlateau'' scheduler with a decay factor of 0.5, a patience of 5 epochs, and maximization mode.

\subsection{Baselines and Evaluation Metric}
To evaluate the performance of our proposed CMTA, we compare it with 13 representative baseline methods. 
According to their specific detection tasks, these baselines are classified into three categories: deepfake video detection, AI-generated image detection, and AI-generated video detection.

We first consider deepfake video detection methods, selecting STIL~\cite{10.1145/3474085.3475508}, FTCN~\cite{zheng2021exploring}, TALL~\cite{xu2023tall}, and MINTIME~\cite{10547206} as representative baselines. In addition, we adapt X-CLIP~\cite{ni2022expanding}, a foundation model for video-text understanding, to our detection task. These methods primarily focus on modeling temporal inconsistencies, learning robust spatiotemporal representations, or leveraging cross-modal alignment to identify sophisticated forgeries.

We also adopt AI-generated image detection approaches, including FID~\cite{NEURIPS2024_6dddcff5} and NPR~\cite{Tan_2024_CVPR}, which detect low-level generative artifacts and CNN-specific upsampling traces within individual frames. To adapt these image-based methods to video detection, video-level predictions are obtained by aggregating frame-level results along the temporal dimension.

Most importantly, we benchmark CMTA against six state-of-the-art methods specifically designed for AI-generated video detection. AIGVDet~\cite{bai2024ai} and DeCoF~\cite{Ma2024DetectingAV} identify synthetic videos by modeling spatiotemporal anomalies and inter-frame inconsistencies, while DeMamba~\cite{chen2024DeMamba} leverages a Mamba-based framework to capture long-range temporal dependencies. Furthermore, we incorporate two physics-driven and geometry-based methods: NSG-VD~\cite{zhang2025NSGVD} and ReStraV~\cite{interno2025aigenerated} detect violations of physical continuity and temporal curvature in the feature space, respectively. Finally, D3~\cite{Zheng_2025_ICCV} introduces a training-free paradigm to uncover motion discrepancies using second-order temporal features.

\emph{Evaluation Metric.} 
To comprehensively evaluate the performance of CMTA, we adopt three widely used metrics for AI-generated video detection: Average Precision (AP), Area Under the ROC Curve (AUC), and Accuracy (ACC).
In line with the evaluation protocols of state-of-the-art methods, AP and AUC serve as the primary metrics for quantifying model discriminability and generalization across diverse decision thresholds, while ACC offers an intuitive measure of overall classification accuracy.

\subsection{Quantitative Results}
Tables~\ref{tab:ap_GenVideo}–\ref{tab:auc_VidProM} present a comprehensive comparison of AP and AUC between CMTA and 13 state-of-the-art methods across four challenging large-scale benchmarks, namely GenVideo, EvalCrafter, VideoPhy, and VidProM.
Meanwhile, Table~\ref{tab:acc_all} compares the classification accuracy (ACC) of CMTA against these baselines on the same benchmark datasets.

\subsubsection{GenVideo}
As shown in Tables~\ref{tab:ap_GenVideo}, \ref{tab:auc_genvideo}, and~\ref{tab:acc_all}, the proposed CMTA establishes a new state-of-the-art on the GenVideo benchmark, consistently outperforming all 13 baseline methods across all metrics.
Specifically, CMTA outperforms the strongest baseline D3 by 6.58\% in AP, and surpasses the second-best baseline ReStraV by 6.61\% in AUC.
In terms of ACC, as reported in Table~\ref{tab:acc_all}, CMTA delivers a substantial improvement of 9.79\% over the previous SOTA competitor DeCoF.
These results underscore the superior discriminative capability and generalization of CMTA in identifying diverse AI-generated videos.

Compared with deepfake video detection methods, CMTA achieves significantly superior performance across all 10 subsets of GenVideo. Unlike conventional deepfake detectors that are inherently designed for face-manipulation forensics and focus on local spatial artifacts and temporal inconsistencies within facial regions, CMTA avoids over-reliance on facial-specific priors and captures more universal generative artifacts beyond localized facial regions. Although such methods effectively capture local frame-level artifacts and short-term temporal inconsistencies, they fail to generalize across diverse video generation models and non-face manipulation videos, resulting in significant performance drops on most subsets. Specifically, STIL and TALL yield 69.68\% and 62.25\% in mean AP, as well as 74.37\% and 60.20\% in mean AUC, respectively. Even the strongest baseline X-CLIP, which benefits from vision-language alignment, is still outperformed by CMTA by a margin of 9.12\% in mean AP. In contrast, CMTA captures holistic video characteristics by constructing cross-modal representations and performing multi-grained temporal modeling, thus avoiding dependence on facial priors and achieving consistent performance across all 10 subsets.

Regarding frame-level AI-generated image detectors (i.e., FID and NPR), these methods are designed to capture low-level generative artifacts and CNN-specific up-sampling traces, and they infer video-level results simply by aggregating frame-wise predictions. However, these frame-wise methods only focus on single-frame spatial artifacts and ignore temporal coherence, resulting in limited performance on generated videos. For example, NPR yields 40.17\% AP on the HotShot subset, while our CMTA maintains 96.99\% AP. This is because NPR exclusively relies on single-frame spatial features and fails to model temporal inconsistencies and unstable dynamics, which are the dominant artifacts in HotShot videos. Similarly, FID performs poorly on the Sora subset, as high-fidelity videos suppress nearly all spatial artifacts, resulting in 74.95\% AP, in contrast to 99.76\% AP achieved by CMTA. Unlike these frame-based methods, CMTA explicitly leverages multi-grained temporal modeling to capture the subtle temporal artifacts that image-level detectors overlook.

In comparison with six dedicated AI-generated video detection methods (i.e., AIGVDet, DeMamba, DeCoF, NSG-VD, ReStraV, and D3), CMTA demonstrates superior performance across the GenVideo benchmark. 
In contrast, existing methods suffer from noticeable limitations when confronted with diverse generators. 
AIGVDet relies on optical flow cues, which become unreliable in high-quality generated videos, yielding 70.46\% mean AP.
DeMamba, built on the Mamba architecture, effectively models long-range dependencies but fails to capture fine-grained forgery traces and performs poorly on high-fidelity subsets such as yielding 28.79\% AP on Sora and 45.07\% AP on Show-1.
Physics-driven methods including NSG-VD and ReStraV depend on handcrafted priors such as gradient statistics or trajectory curvature, which struggle to adapt to complex temporal variations and diverse generative models in GenVideo.
Notably, in extremely challenging scenarios like Sora, where most baselines fall below 70\% AP, CMTA maintains near-perfect performance at 99.76\% AP.
Such superior performance stems from CMTA’s unique dual advantages: i) the cross-modal fusion of visual semantics and textual descriptions greatly enhances feature discriminability; 
ii) multi-grained temporal modeling effectively captures full-scale artifacts ranging from short-range local inconsistencies to long-range global motion anomalies. This validates the effectiveness and strong generalization ability of its multi-modal and multi-grained modeling design.

\subsubsection{EvalCrafter}
Tables~\ref{tab:ap_EvalCrafter}, \ref{tab:auc_EvalCrafter}, and~\ref{tab:acc_all} present a detailed performance comparison between CMTA and 13 baseline methods on the EvalCrafter dataset.
Comprising 14 distinct subsets from diverse generative models and their variants (e.g., Gen2, LaVie, and PikaLab), EvalCrafter provides a rigorous setting to evaluate the generalization of detectors across various model architectures, resolutions, and temporal smoothness levels.
Collectively, CMTA achieves a new state-of-the-art on this benchmark, outperforming the most competitive baseline, FID, by 4.14\% in mean AP and 4.31\% in mean AUC. 
Furthermore, CMTA delivers a substantial accuracy improvement of 8.51\% over the previous state-of-the-art method DeCoF. 
These results underscore CMTA's stable discriminative capability against diverse generators, training paradigms, and visual quality levels.

Existing approaches exhibit certain limitations when evaluated across the diverse scenarios in EvalCrafter.
Deepfake detection methods such as STIL and TALL, which rely on facial-specific priors, suffer from significant performance drops in non-face and high-motion samples. 
Similarly, frame-level image detectors like FID and NPR, despite performing well on subsets with static spatial artifacts, fail to maintain stability in motion-intensive scenarios like HotShot-XL, where low-level spatial traces are suppressed. 
Regarding dedicated AI-generated video detectors, AIGVDet and DeMamba exhibit varying degrees of degradation across subsets, particularly on the LaVie-Base and LaVie-Interpolation variants. 
This highlights their limited generalization against subtle temporal artifacts and unstable motion patterns inherent in different model versions.

In contrast, CMTA maintains consistently high performance across all 14 subsets, demonstrating exceptional generalization to both intense motion dynamics and subtle temporal inconsistencies. 
These results validate that CMTA does not rely on specific generative biases but instead captures universal AI-generated artifacts, ensuring strong generalization across evolving model versions and varying video quality.

\begin{table*}[!t]
    \centering
    \caption{Ablation results of CMTA on GenVideo, evaluating image captioning backbones and visual-textual encoders in terms of AP (\%).}
    \resizebox{\textwidth}{!}{
    \begin{tabular}{c c | c c c c c c c c c c | c}
    \toprule
    Caption Model              & Visual \& Textual Encoder    & MS       & MPS       & MV       & HotShot       & Show-1       & Gen2       & Crafter       & LaVie       & Sora       & WS       & mean   \\
    \midrule
    \hline
    \multirow{4}{*}{BLIP-large}     
                               & XCLIP-P16           & 97.84    & 98.02     & 99.35    & 95.40         & 96.45        & 98.50      & 98.42         & \underline{95.66}       & 97.27      & \underline{93.30}    & 97.02   \\ 
                               & XCLIP-P32           & 93.24    & 94.39     & 96.12    & 89.75         & 91.26        & 96.63      & 93.79         & 92.06       & 94.82      & 88.32    & 93.04   \\ 
                               & CLIP-P16        & 97.50    & 97.79     & \textbf{99.71}    & 84.37         & 96.23        & \underline{99.39}      & 98.57         & 90.72       & 99.65      & 89.31    & 95.32   \\ 
                               & CLIP-P32        & 97.36    & 96.72     & 98.47    & 91.77         & 93.12        & 97.93      & 95.86         & 88.70       & \textbf{100.00}     & 90.30    & 95.02   \\
    \hline
    \multirow{5}{*}{BLIP-base}   
                               & XCLIP-P16           & 97.98         & 97.28          & 98.93         & \underline{95.63}              & \underline{96.65}             & 98.88           & 97.85              & \underline{95.66}            & 99.52           & 92.80         & \underline{97.12}   \\ 
                               & XCLIP-P32           & 91.84         & 93.79          & 96.63         & 88.01              & 89.91             & 97.56           & 95.33              & 91.13            & 97.72           & 90.07         & 93.20   \\  
                               & CLIP-P16        & \underline{98.77}         & \underline{98.76}          & \underline{99.60}         & 84.79              & 95.38             & \underline{99.39}           & \underline{98.97}              & 92.86            & 98.35           & 90.45         & 95.73   \\ 
                               & \cellcolor{gray!25}CLIP-P32        & \cellcolor{gray!25}\textbf{99.22}     & \cellcolor{gray!25}\textbf{99.59}    & \cellcolor{gray!25}96.09              & \cellcolor{gray!25}\textbf{96.99}     & \cellcolor{gray!25}\textbf{97.65}    & \cellcolor{gray!25}\textbf{99.98}     & \cellcolor{gray!25}\textbf{99.69}        & \cellcolor{gray!25}\textbf{99.29}      & \cellcolor{gray!25}\underline{99.76}    & \cellcolor{gray!25}\textbf{99.10}         & \cellcolor{gray!25}\textbf{98.74}      \\
    \bottomrule
    \end{tabular}
    }
\label{tab:ablation2_ap}
\end{table*}

\begin{table*}[ht]
    \centering
    \caption{Ablation results of CMTA on GenVideo, evaluating image captioning backbones and visual-textual encoders in terms of AUC (\%).}
    \resizebox{\textwidth}{!}{
    \begin{tabular}{c c | c c c c c c c c c c | c}
    \toprule
    Caption Model              & Visual \& Textual Encoder    & MS       & MPS       & MV       & HotShot       & Show-1       & Gen2       & Crafter       & LaVie       & Sora       & WS       & mean   \\
    \midrule
    \hline
    \multirow{5}{*}{BLIP-large}  
                               & XCLIP-P16           & 97.96    & 98.15     & 99.32    & \underline{95.61}         & 96.64        & 98.62      & 98.37         & \underline{95.79}       & 96.94      & \underline{92.20}    & 96.96   \\ 
                               & XCLIP-P32           & 93.61    & 94.38     & 96.60    & 90.25         & 91.76        & 96.95      & 94.03         & 92.01       & 94.13      & 87.23    & 93.10   \\    
                               & CLIP-P16        & 97.48         & 98.00          & \textbf{99.71}         & 83.91              & 96.32             & \underline{99.36}           & 98.45              & 89.97            & 99.62           & 86.96         & 94.98   \\ 
                               & CLIP-P32        & 97.58    & 97.18     & 98.55    & 91.69         & 94.19        & 98.41      & 96.49         & 89.07       & \textbf{100.00}     & 88.44    & 95.16   \\
    \hline 
    \multirow{5}{*}{BLIP-base}    
                               & XCLIP-P16           & 97.99    & 97.46     & 98.86    & 95.38         & \underline{96.74}        & 98.89      & 97.80         & 95.50       & 99.49      & 91.65    & \underline{96.98}   \\ 
                               & XCLIP-P32           & 92.37    & 93.58     & 96.88    & 90.06         & 89.74        & 97.52      & 95.28         & 91.51       & 97.32      & 88.21    & 93.25   \\ 
                               & CLIP-P16        & \underline{98.65}    & \underline{98.68}     & \underline{99.57}    & 84.29         & 95.66        & 99.33      & \underline{98.94}         & 91.52       & 96.94      & 87.71    & 95.13   \\
                               & \cellcolor{gray!25}CLIP-P32        & \cellcolor{gray!25}\textbf{99.41} & \cellcolor{gray!25}\textbf{99.78}  & \cellcolor{gray!25}97.92  & \cellcolor{gray!25}\textbf{98.31}  & \cellcolor{gray!25}\textbf{98.64}    & \cellcolor{gray!25}\textbf{99.98}    & \cellcolor{gray!25}\textbf{99.51} & \cellcolor{gray!25}\textbf{98.96} & \cellcolor{gray!25}\underline{99.74}    & \cellcolor{gray!25}\textbf{98.79}          & \cellcolor{gray!25}\textbf{99.10}  \\ 
    \bottomrule
    \end{tabular}
    }
\label{tab:ablation2_auc}
\end{table*}

\begin{table*}[ht]
    \centering
    \caption{Ablation results of CMTA on GenVideo, evaluating image captioning backbones and visual-textual encoders in terms of ACC (\%).}
    \resizebox{\textwidth}{!}{
    \begin{tabular}{c c | c c c c c c c c c c | c}
    \toprule
    Caption Model              & Visual \& Textual Encoder    & MS       & MPS       & MV       & HotShot       & Show-1       & Gen2       & Crafter       & LaVie       & Sora       & WS       & mean   \\
    \midrule
    \hline
    \multirow{5}{*}{BLIP-large}  
                               & XCLIP-P16           & 91.43    & 91.57     & 93.45    & \underline{88.43}         & \underline{90.00}        & 91.81      & 90.92         & \underline{88.50}       & 92.86      & \underline{84.60}    & 90.36   \\ 
                               & XCLIP-P32           & 85.71    & 86.00     & 89.62    & 82.57         & 83.57        & 89.86      & 85.91         & 84.43       & 85.71      & 78.91    & 85.23   \\   
                               & CLIP-P16        & 90.14    & 90.43     & 91.85    & 74.86         & 87.86        & 92.03      & 89.06         & 82.07       & 92.86      & 79.02    & 87.02   \\ 
                               & CLIP-P32        & 90.43    & 92.00     & \textbf{94.73}    & 80.00         & 84.71        & 94.20      & 91.42         & 78.00       & \textbf{98.21}      & 79.24    & 88.29   \\ 
    \hline  
    \multirow{5}{*}{BLIP-base}   
                               & XCLIP-P16           & 92.14    & 91.14     & \underline{94.41}    & 87.43         & 89.71        & \underline{94.64}      & 92.20         & 87.36       & \underline{96.43}      & 83.93    & \underline{90.94}   \\ 
                               & XCLIP-P32           & 83.71    & 85.14     & 89.94    & 82.57         & 80.14        & 91.38      & 87.98         & 83.00       & 91.07      & 79.80    & 85.47   \\  
                               & CLIP-P16        & \underline{92.86}    & \underline{93.00}     & 93.93    & 75.14         & 89.57        & 93.48      & \underline{92.92}         & 84.14       & \underline{96.43}      & 81.14    & 89.26   \\ 
                               & \cellcolor{gray!25}CLIP-P32        & \cellcolor{gray!25}\textbf{98.29} & \cellcolor{gray!25}\textbf{99.71}  & \cellcolor{gray!25}94.09 & \cellcolor{gray!25}\textbf{97.86}  & \cellcolor{gray!25}\textbf{98.14}  & \cellcolor{gray!25}\textbf{99.49}  & \cellcolor{gray!25}\textbf{98.93}   & \cellcolor{gray!25}\textbf{97.14}   & \cellcolor{gray!25}\underline{96.43}    & \cellcolor{gray!25}\textbf{93.86}     & \cellcolor{gray!25}\textbf{97.39}  \\
    \hline
    \bottomrule
    \end{tabular}
    }
\label{tab:ablation2_acc}
\end{table*}

\subsubsection{VideoPhy}
Tables~\ref{tab:ap_VideoPhy}, \ref{tab:auc_VideoPhy}, and~\ref{tab:acc_all} compare CMTA with 13 baselines on the VideoPhy benchmark.
Unlike general AI-generated video detection datasets, VideoPhy is uniquely designed to evaluate physical plausibility violations, including unnatural rigid-body deformation, inconsistent fluid dynamics, non-physical penetration, and violations of basic physical laws.
This setting provides a challenging testbed to assess whether detectors can capture high-level inconsistencies in motion dynamics, rather than only low-level visual artifacts.
CMTA sets a new state-of-the-art on VideoPhy, with a mean AP of 95.54\% and mean AUC of 97.34\%, surpassing all baselines by clear margins.
It also achieves a mean accuracy of 93.66\%, demonstrating a significant improvement over existing competitors.
Specifically, CMTA outperforms the strongest baseline FID by 0.85\% in AP and 3.01\% in AUC, and exceeds the second-best method MINTIME by 16.53\% in accuracy.
These results demonstrate CMTA’s strong capability to identify violations in physical consistency, especially its notable advantage in AUC, which indicates reliable discrimination of unnatural and non-physical motion patterns.

Deepfake detection methods such as STIL and TALL perform the worst on VideoPhy, with mean AP of 68.73\% and 51.98\% respectively. This is because they are inherently designed for facial manipulation detection, and completely lack the ability to model general motion dynamics and physical rules.
Image-level detectors like FID and NPR achieve competitive overall performance by capturing low-level visual traces, but they suffer severe performance drops on fluid-dominated subsets. 
For instance, NPR drops sharply to 63.72\% AP and 42.90\% AUC on LaVie, because it relies on single-frame visual features and cannot effectively model temporal consistency in fluid motions.
Among physics-aware baselines, ReStraV and D3 demonstrate strong performance on specific subsets. For example, ReStraV achieves 100.00\% AP on Pika and 99.65\% AUC on OpenSora, while D3 obtains the highest AP and AUC on the CVX-5B subset.
Specifically, ReStraV drops to 61.39\% AP and 65.21\% AUC on CVX-5B, and D3 struggles with fluid-dominated sequences, such as LaVie, where it performs worse than CMTA by a large margin of 15.62\% in AP.
In contrast, CMTA maintains consistently high performance across all 10 subsets, including the most challenging fluid and multi-body interaction cases. 
For example, CMTA achieves 94.28\% AP on CVX and 97.84\% AP on LaVie, outperforming physics-driven methods like D3 and remaining highly competitive with ReStraV in these complex physical scenarios.
This indicates that CMTA implicitly captures object motion dynamics via multi-grained temporal modeling and visual-textual alignment, instead of relying on predefined physical rules, leading to stronger generalization across diverse physical violation scenarios.

Overall, the results on VideoPhy validate that CMTA enables effective identification of high-level physical plausibility violations, going beyond the limitations of low-level artifact detection and handcrafted physical prior approaches.

\subsubsection{VidProM}
As illustrated in Tables~\ref{tab:ap_VidProM}, \ref{tab:auc_VidProM}, and~\ref{tab:acc_all}, CMTA achieves the best overall performance on VidProM across all metrics.
Notably, VidProM consists of videos generated from real user prompts, which are typically longer, more complex, and involve open-ended scenarios, making it a critical benchmark for evaluating real-world generalization.
Specifically, CMTA obtains a mean AP of 94.35\% and a mean AUC of 95.86\%, outperforming the second-best method DeCoF by 4.24\% and 5.09\%, respectively. 
In terms of accuracy, CMTA reaches 89.35\%, surpassing the closest competitor DeCoF by a margin of 3.88\%.

Compared with deepfake detection methods, CMTA demonstrates substantial superiority in handling the rich semantic diversity induced by complex prompts.
Specifically, STIL achieves 61.29\% in mean AP and 65.54\% in mean AUC, while TALL attains a lower mean AP of 56.85\% and a mean AUC of 55.03\%.
These results indicate that methods relying on local facial priors struggle to generalize to the open-domain content distribution of VidProM.
Similarly, image-level artifact detectors such as FID and NPR exhibit limited adaptability to these complex semantic scenarios.
Although FID achieves a high AP of 97.87\% on the ST2V subset, its performance drops significantly to 68.51\% on T2VZ. This sharp decline reveals its inherent limitation in maintaining stability across diverse generation paradigms.
Regarding specialized AI-generated video detection methods, they demonstrate weak generalization on the diverse content distribution present in VidProM.
For instance, ReStraV yields a mean AP of 77.80\% and a mean AUC of 77.70\%, failing to handle the variety of generative models.
Furthermore, AIGVDet suffers from severe performance degradation, with its mean AP and AUC dropping to 60.44\% and 49.07\%, respectively. 
Its performance is particularly low on typical subsets including VC2 and ST2V, which further demonstrates its limited generalization ability.

In contrast, CMTA leverages multi-grained temporal modeling to discern subtle cross-modal discrepancies, making it inherently suited to processing AI-generated videos conditioned on complex prompts. 
This allows it to capture effective global semantic structures and maintain consistent, high-performance results across all subsets.
It obtains the best AUC on T2VZ and VC2, and achieves the highest mean AP and mean AUC among all methods on the VidProM benchmark.
These results on VidProM validate that CMTA possesses strong practical deployment potential in real-world scenarios characterized by open-ended prompts and diverse semantics.

\subsection{Ablation Studies}
To validate the effectiveness of each key component in CMTA and investigate the impact of different pre-trained backbones, we conduct comprehensive ablation studies on the GenVideo benchmark.
All experiments are conducted under the same settings, and performance is evaluated using AP, AUC, and ACC for comprehensive comparison.

\subsubsection{Impact of Key Components}
To quantify the individual contribution of each component in CMTA, we conduct ablation studies on the GenVideo benchmark with four variants:
(i) V-only, which employs only visual representations; 
(ii) T-only, which relies exclusively on textual semantics;
(iii) VT-CGTM, which combines cross-modal features with coarse-grained temporal modeling;
and (iv) VT-FGTM, which leverages visual-textual representations equipped with fine-grained temporal modeling.
The corresponding results in terms of AP, AUC, and ACC are presented in Tables~\ref{tab:ablation_ap}, ~\ref{tab:ablation_auc}, and~\ref{tab:ablation_acc}, respectively.

Experimental results show that the V-only variant achieves the most competitive performance among all ablated variants, with reductions of 5.05\%, 4.97\%, and 11.77\% in AP, AUC, and ACC, respectively, compared with the full CMTA model.
This gap reveals that although visual features alone capture critical synthetic artifacts, they cannot sufficiently exploit the discriminative cross-modal temporal patterns facilitated by joint visual-textual modeling.
Conversely, the T-only variant exhibits the most inferior performance, with its AP and AUC dropping by more than 30 percentage points.
This demonstrates that while high-level textual descriptions provide contextual semantics, they lack the fine-grained spatial details and low-level artifacts that are indispensable for identifying subtle AI-generated traces.
Notably, both VT-CGTM and VT-FGTM are inferior to the V-only baseline, which potentially stems from the semantic interference caused by incomplete cross-modal temporal modeling.
Without dual-grained temporal modeling, textual features may introduce noise that disrupts the visual feature space, rather than facilitating effective multi-modal fusion, thereby degrading generalization and stability.
Specifically, VT-CGTM undergoes substantial performance degradation, with declines exceeding 20 percentage points in all metrics, whereas VT-FGTM demonstrates moderate performance drops, with reductions of 7.92\% in AP, 8.35\% in AUC, and 16.07\% in ACC.

In contrast, the full CMTA framework effectively harmonizes visual and textual modalities via dual-grained temporal modeling, transforming textual semantics from potential interference into complementary semantic cues.
Consequently, the full CMTA framework delivers superior generalization compared to its ablated variants.
These results confirm that combining coarse-grained and fine-grained temporal modeling is essential for capturing discriminative cross-modal temporal artifacts, 
thereby maintaining reliable detection performance across complex generative scenarios.

\begin{figure*}[!ht]
    \centering
    \subfloat[{\footnotesize Crafter}]{\includegraphics[width=0.19\textwidth]{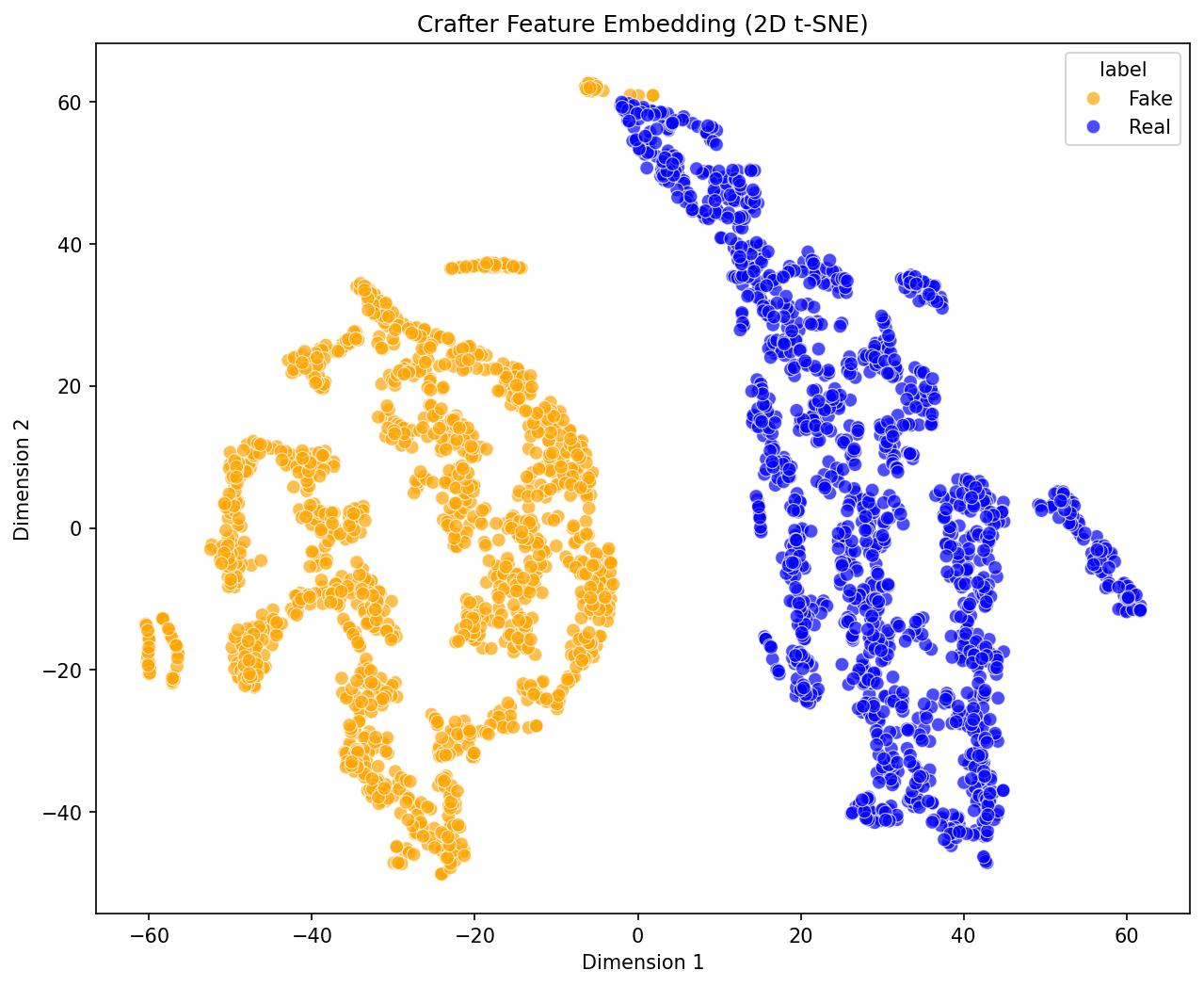}}\hfill
    \subfloat[{\footnotesize Gen2}]{\includegraphics[width=0.19\textwidth]{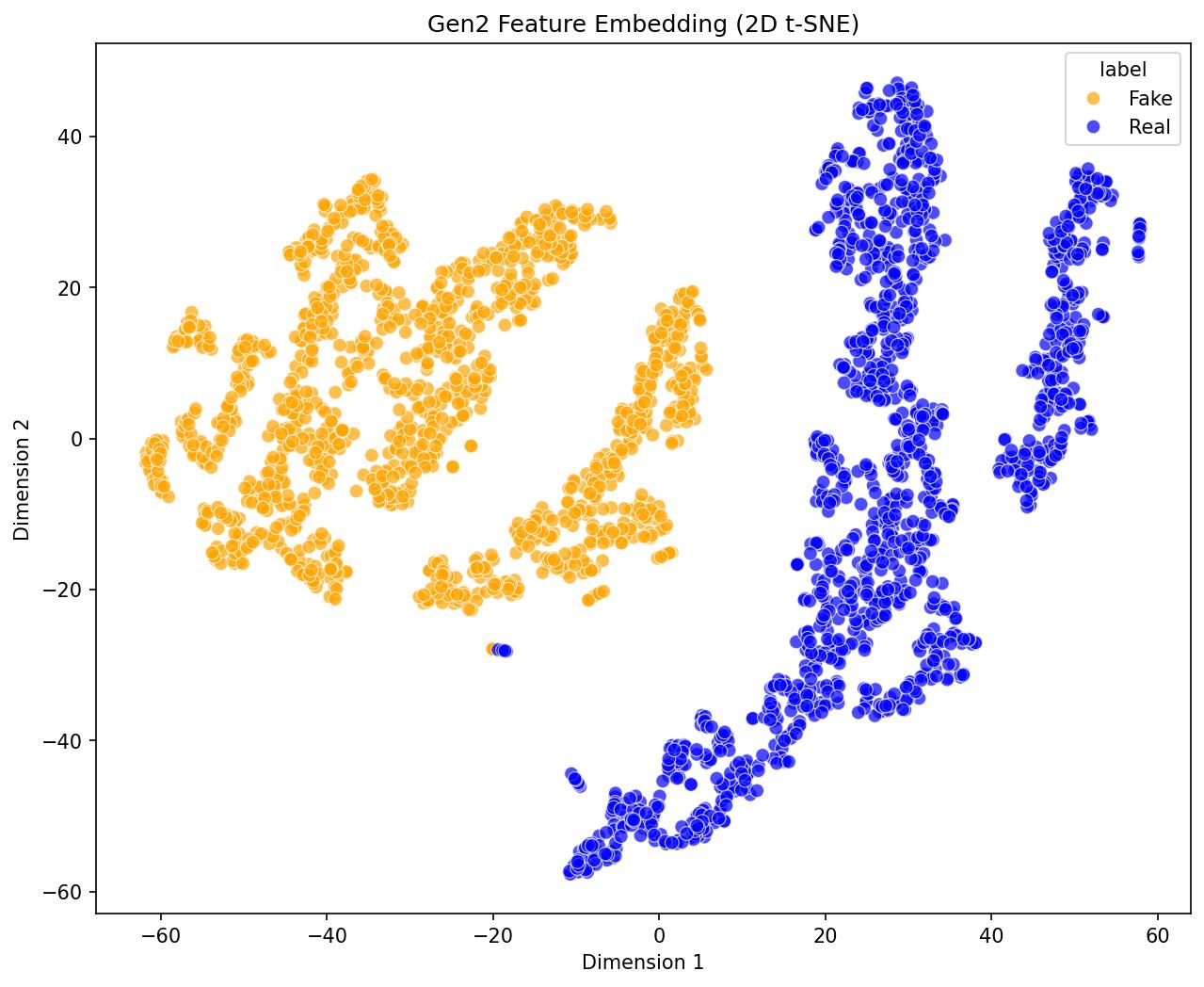}}\hfill
    \subfloat[{\footnotesize HotShot}]{\includegraphics[width=0.19\textwidth]{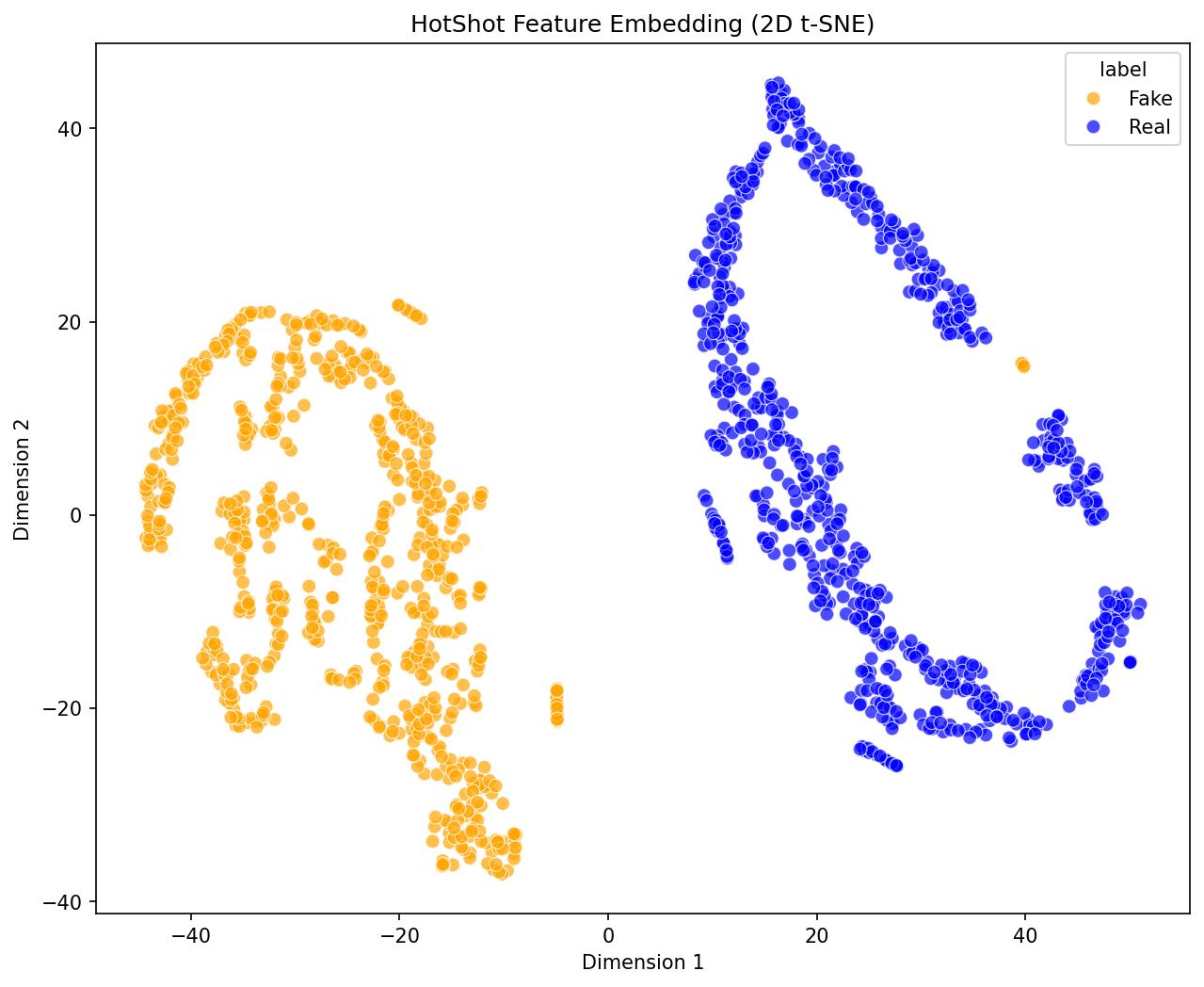}}\hfill
    \subfloat[{\footnotesize Lavie}]{\includegraphics[width=0.19\textwidth]{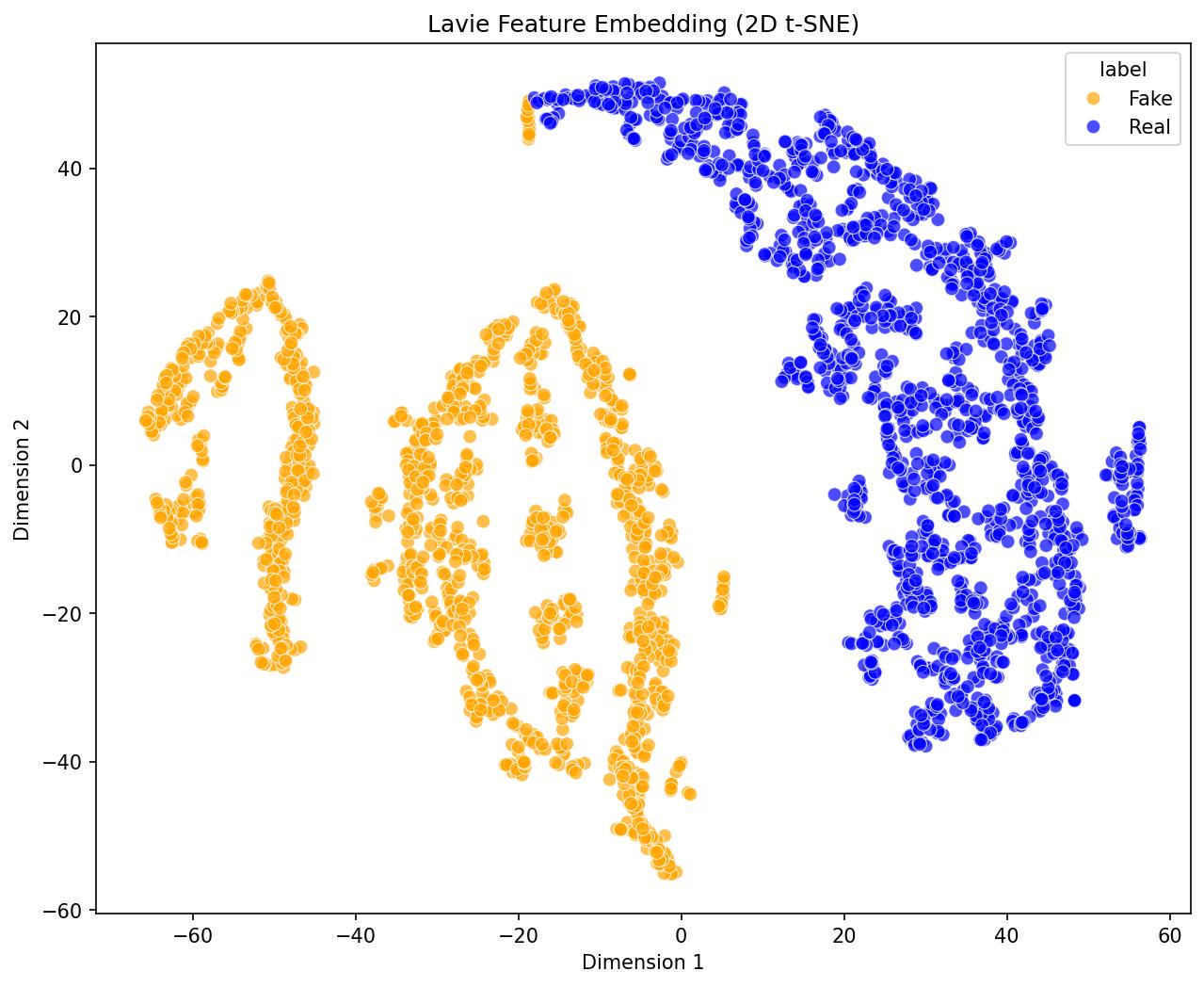}}\hfill
    \subfloat[{\footnotesize ModelScope}]{\includegraphics[width=0.19\textwidth]{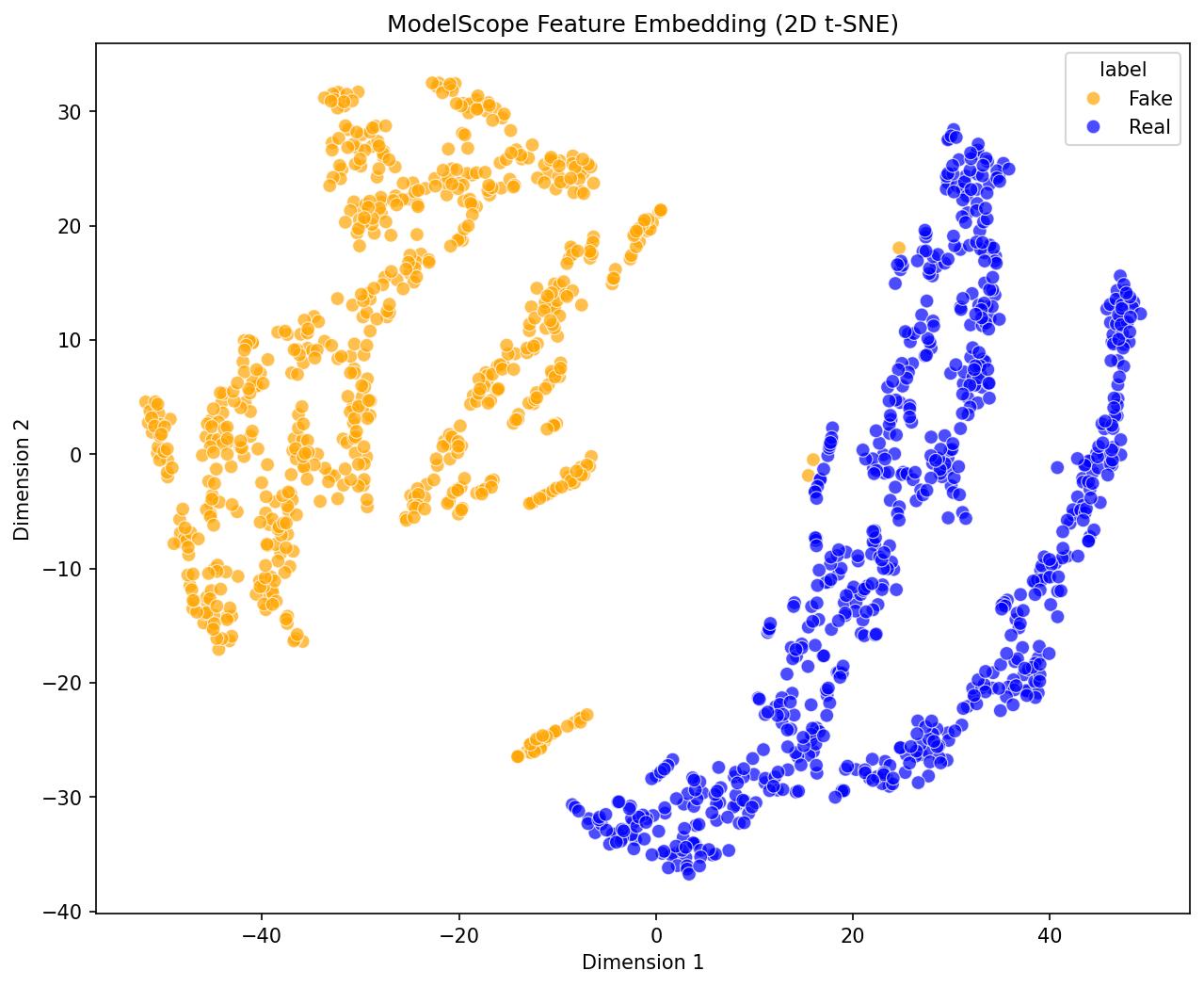}}\\[-0.2em]
    \subfloat[{\footnotesize MoonValley}]{\includegraphics[width=0.19\textwidth]{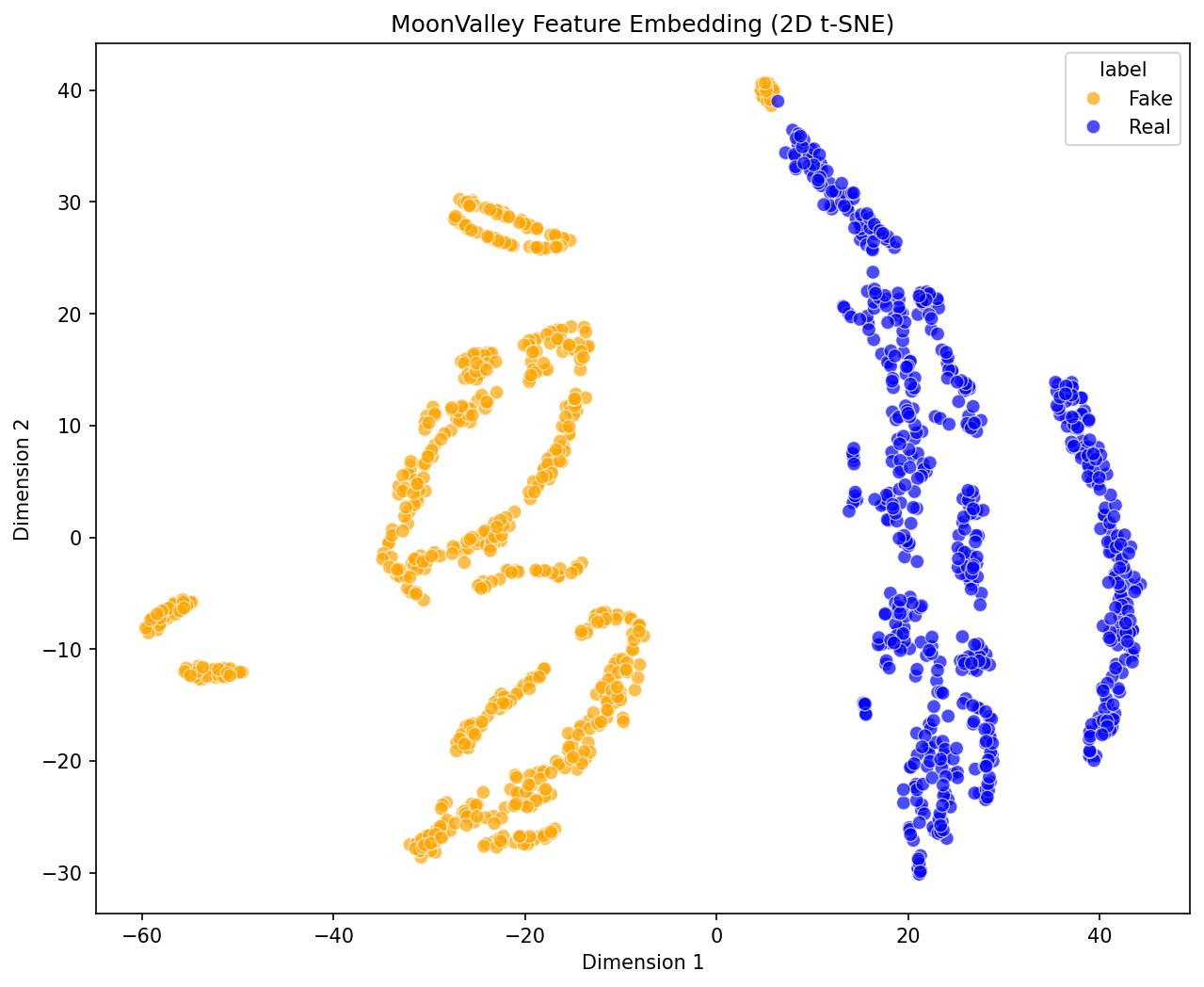}}\hfill
    \subfloat[{\footnotesize MorphStudio}]{\includegraphics[width=0.19\textwidth]{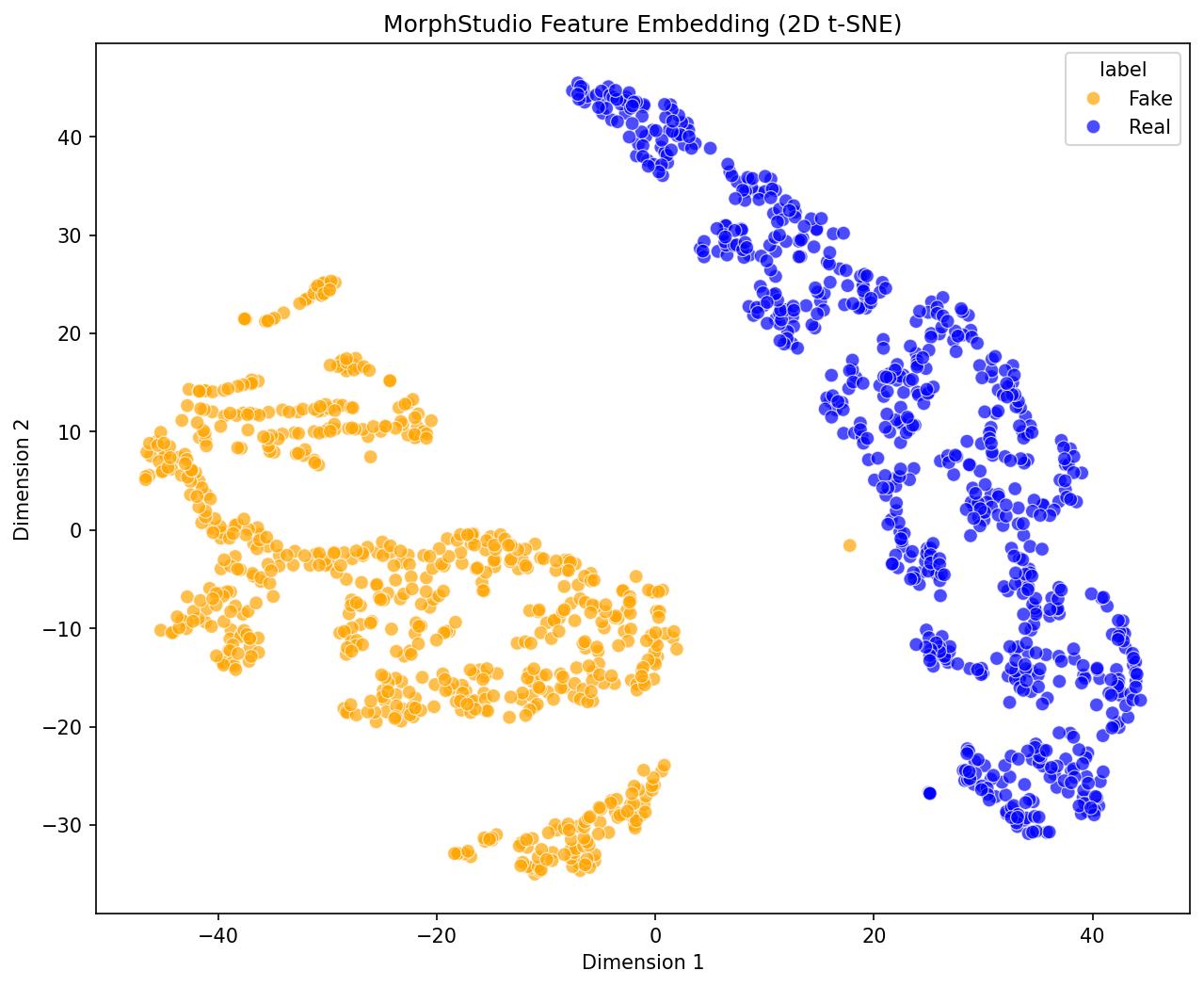}}\hfill
    \subfloat[{\footnotesize Show-1}]{\includegraphics[width=0.19\textwidth]{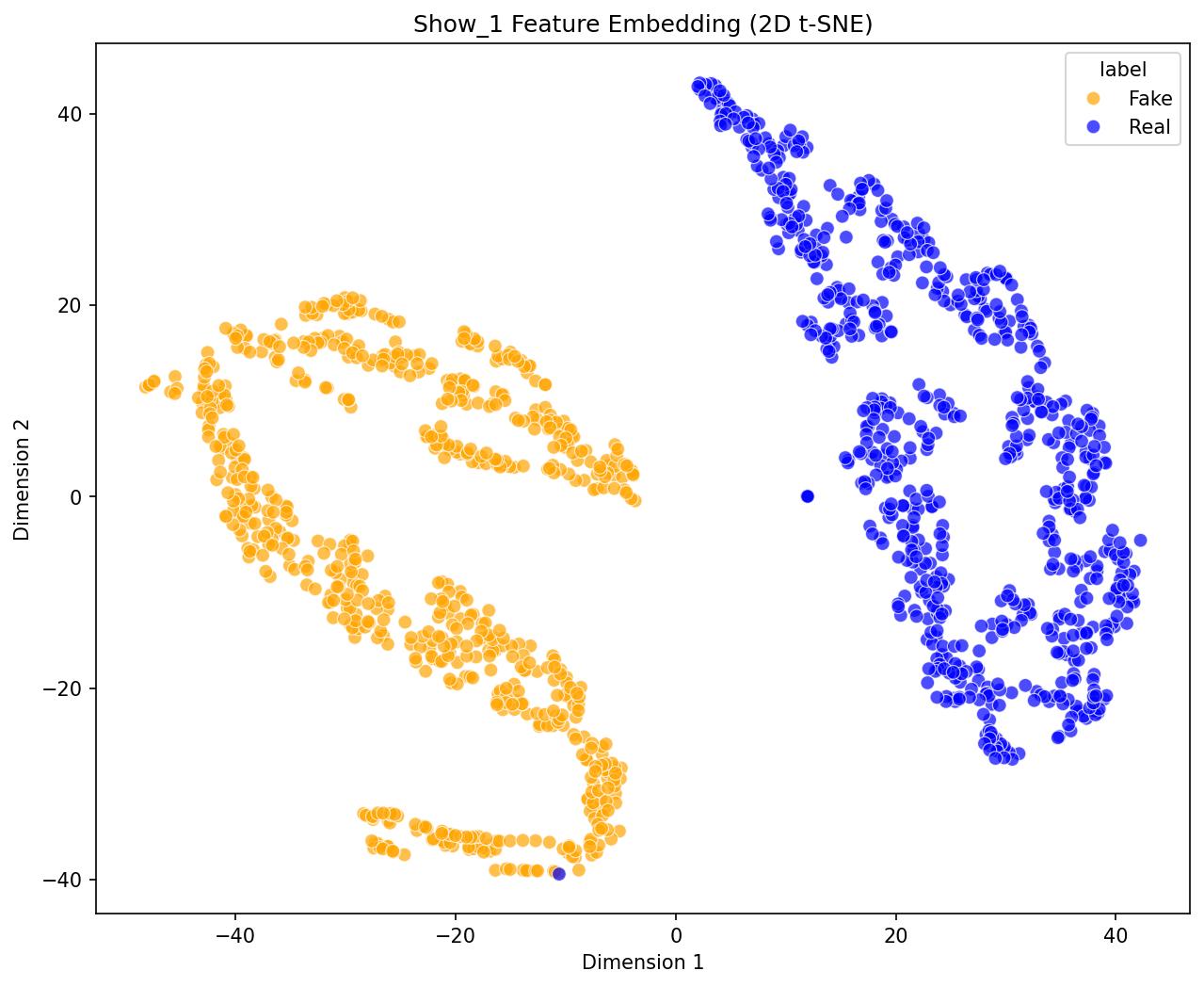}}\hfill
    \subfloat[{\footnotesize Sora}]{\includegraphics[width=0.19\textwidth]{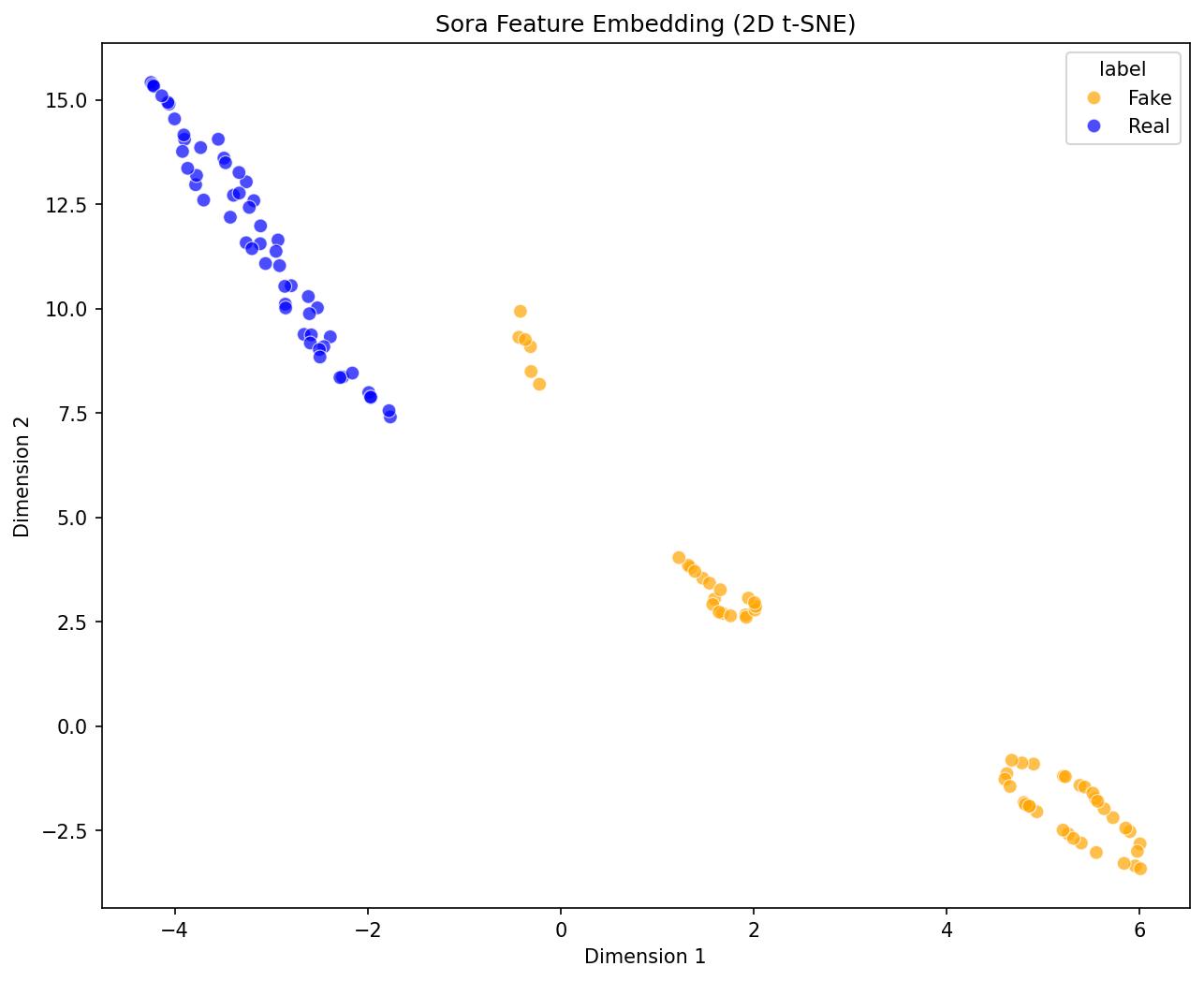}}\hfill
    \subfloat[{\footnotesize WildScrape}]{\includegraphics[width=0.19\textwidth]{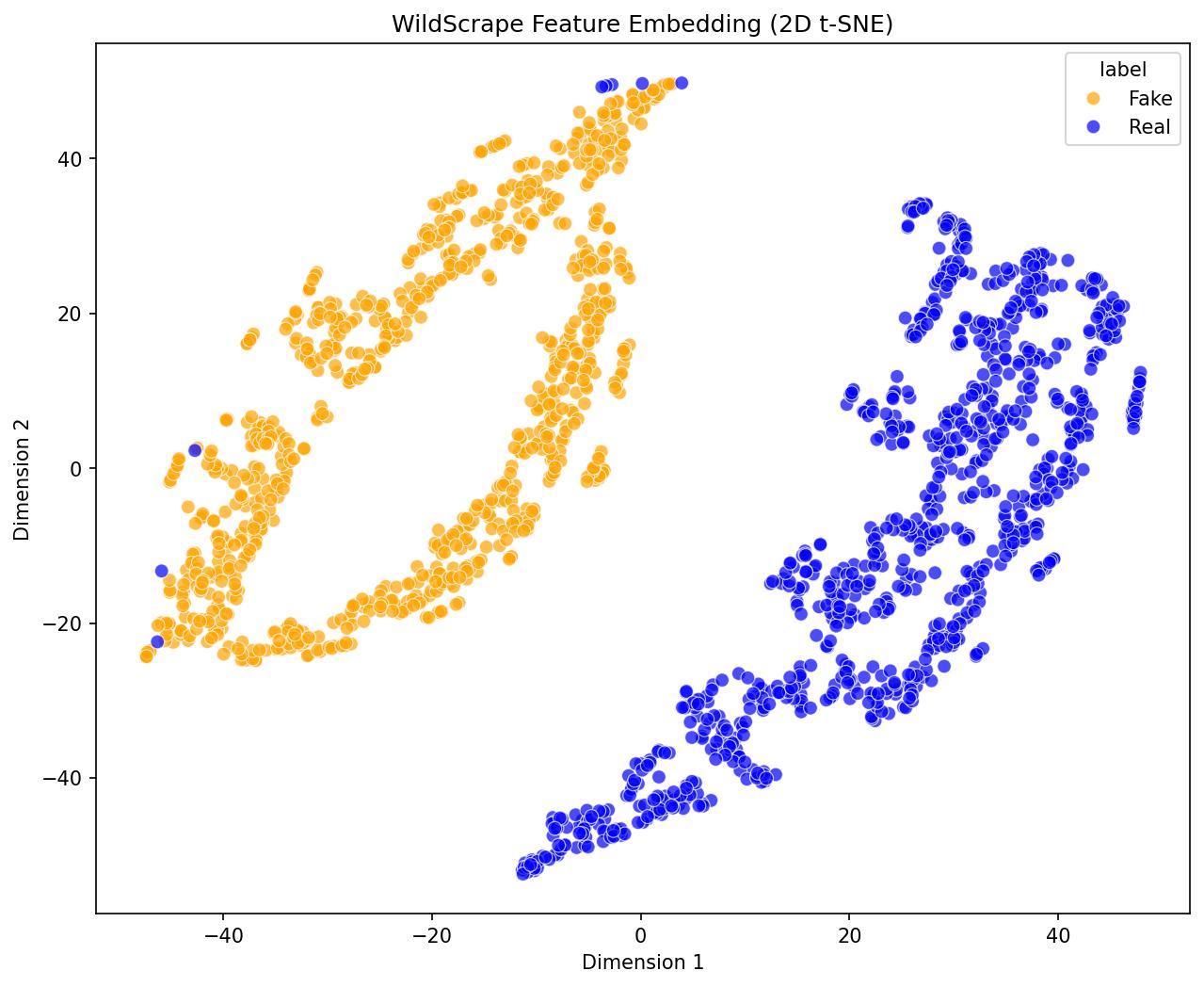}}
    \caption{t-SNE visualization of features extracted before the classification head. Blue and orange points represent real and AI-generated videos, respectively. We visualize features on 10 AI-generated video subsets from the GenVideo test set, where real videos are randomly sampled from MSR-VTT with a 1:1 balanced ratio for each subset.}
    \label{fig:tsne_visualization}
\end{figure*}

\subsubsection{Impact of Pre-trained Backbones}
To investigate the impact of pre-trained backbones on CMTA, we conduct a comprehensive evaluation across different combinations of image captioning models and visual-textual encoders.
Specifically, we employ two image captioning models, ``blip-image-captioning-base'' (BLIP-base) and ``blip-image-captioning-large'' (BLIP-large), paired with four representative encoders: ``XCLIP-base-patch16'' (XCLIP-P16), ``XCLIP-base-patch32'' (XCLIP-P32), ``CLIP-ViT-base-patch16'' (CLIP-P16), and ``CLIP-ViT-base-patch32'' (CLIP-P32). 
The corresponding results, evaluated in terms of AP, AUC and ACC, are reported in Tables~\ref{tab:ablation2_ap}--\ref{tab:ablation2_acc}.

Compared with BLIP-large, BLIP-base consistently yields superior results across all metrics and encoders, with improvements of 3.72\% in mean AP, 3.94\% in mean AUC, and 9.10\% in mean ACC when paired with CLIP-P32. 
This indicates that the BLIP-base backbone provides more effective semantic priors for uncovering cross-modal temporal artifacts.
In contrast, while the BLIP-large model generates more detailed descriptive captions, it may introduce excessive linguistic complexity and semantic redundancies that complicate cross-modal fusion, thereby degrading overall discriminative performance.

Further ablations on the visual-textual encoders suggest that the CLIP-P32 encoder is the optimal choice for CMTA. 
When combined with BLIP-base, it achieves the highest mean scores of 98.74\% AP, 99.10\% AUC, and 97.39\% ACC, outperforming other encoder combinations by clear margins. 
Specifically, it surpasses XCLIP-P16 by 1.62\% AP, 2.12\% AUC, and 6.45\% ACC, underscoring its stronger capacity to model global visual-textual dependencies and capture discriminative temporal patterns.
Interestingly, the performance of XCLIP-P32 variants is markedly inferior to that of XCLIP-P16 counterparts.
The underlying reason may be that XCLIP is adept at capturing local, high-resolution details, and larger patch sizes induce a granularity mismatch that impairs cross-modal alignment, thereby hindering the model's efficacy in identifying complex generative artifacts.

Notably, the optimal configuration (i.e., BLIP-base + CLIP-P32) consistently delivers leading performance on the majority of subsets, including challenging scenarios like HotShot, LaVie, and WS, thereby underscoring its superior generalization capability.
This combination achieves the best balance between semantic richness and feature stability, facilitating reliable detection of cross-modal temporal artifacts in diverse generative scenarios.

\subsection{Qualitative Results}
As illustrated in Fig.~\ref{fig:tsne_visualization}, we utilize t-SNE to project the fused representations into a 2D space for visualization.
The samples consist of AI-generated videos from ten GenVideo test subsets and real videos randomly sampled from MSR-VTT, with a balanced 1:1 real-to-AI ratio.
Across all subsets, we observe two well-separated clusters with large inter-class margins, indicating that the learned representation is nearly linearly separable.
Real videos typically form elongated, arc-like manifolds, which suggests higher diversity in content and motion patterns.
In contrast, AI-generated videos usually split into multiple compact modes, consistent with generator- and sampling-specific artifacts that the model effectively captures.
Specifically, \emph{Crafter} and \emph{HotShot} show particularly clean bipartite structures, with almost no overlap between the two classes.
\emph{Gen2}, \emph{Lavie}, and \emph{MoonValley} exhibit multi-modal distributions for AI-generated videos, while real videos span a broader manifold; yet the inter-class separation remains consistently wide.
\emph{ModelScope} presents a few generated samples near the boundary of the real cluster, potentially reflecting compression artifacts or out-of-distribution content.
\emph{MorphStudio} yields a compact, disk-like generated cluster, opposite a distinct arc-shaped real cluster.
\emph{Show-1} forms an S-shaped generated cluster, with negligible cross-cluster outliers.
Despite having fewer samples, \emph{Sora} remains cleanly separated. 
Finally, \emph{WildScrape} shows two nearly parallel arcs, suggesting that the model relies on global statistics rather than content-specific cues to distinguish real and generated videos.

\section{Conclusion}
\label{sec:concl}
In this paper, we presented CMTA, a novel cross-modal temporal framework designed to detect AI-generated videos. 
CMTA first generated frame-level image captions and extracted corresponding visual-textual representations using pre-trained vision-language models. 
Then it employed a GRU to model coarse-grained temporal fluctuations in cross-modal similarity and a Transformer encoder to capture fine-grained temporal variations in frame-wise visual-textual features. 
By fusing these multi-grained cross-modal temporal clues, CMTA effectively captured the abnormally stable cross-modal alignment patterns that are typical of AI-generated videos and are often overlooked by uni-modal methods. 
Extensive experiments on 40 subsets across four large-scale benchmarks, including GenVideo, EvalCrafter, VideoPhy, and VidProM, demonstrated that CMTA achieved state-of-the-art performance in terms of AP, AUC, and ACC, while delivering strong generalization across various video generators.

\section*{Acknowledgments}
This research was partially supported by the National Natural Science Foundation of China (62441238, U24B20185).
The Gemini 3 model was employed to assist in linguistic polishing and improving the readability of this manuscript. 
Specifically, this AI tool was used exclusively for language refinement and did not contribute to research conception, methodology, experimental analysis, or the formulation of scientific conclusions. 
The authors bear full responsibility for the final content and ensure that all revised text adheres to ethical guidelines, remaining free from plagiarism or scientific misconduct.

\bibliographystyle{IEEEtran}
\bibliography{ref}

\vfill

\end{document}